\documentclass[10pt,twocolumn,letterpaper]{article}

\usepackage{iccv}
\usepackage{times}
\usepackage{epsfig}
\usepackage{graphicx}
\usepackage{amsmath}
\usepackage{amssymb}

\usepackage{subfigure}
\usepackage{amsthm}
\usepackage{enumitem}

\usepackage[toc,page]{appendix}

\iccvfinalcopy % *** Uncomment this line for the final submission

\def\ie{i.e.}
\def\eg{e.g.}
\def\st{\textrm{s.t.}}
\def\and{\textrm{and}}

\def\aff{\textrm{aff}}
\def\dim{\textrm{dim}}
\def\diag{\textrm{diag}}

\def\card{\textrm{card}}

\def\spann{\textrm{span}}

\def\0{\boldsymbol{0}}
\def\1{\textbf{1}}
\def\c{\boldsymbol{c}}

\def\v{\boldsymbol{v}}
\def\p{\boldsymbol{p}}

\def\w{\boldsymbol{w}}
\def\x{\boldsymbol{x}}

\def\II{\mathcal{I}}
\def\I{\mathbf{I}}

\def\C{\mathbf{C}}
\def\P{\mathbf{P}}

\def\X{\mathbf{X}}
\def\W{\mathbf{W}}

\def\cA{\mathcal{A}}

\def\cH{\mathcal{H}}
\def\cT{\mathcal{T}}
\def\cS{\mathcal{S}}
\def\cW{\mathcal{W}}
\def\cX{\mathcal{X}}

\def\transpose{\top} % Vector and Matrix Transpose
\def\st{\hspace{1em} \mathrm{s.t.} \hspace{0.5em}}

\newcommand{\RR}{I\!\!R} %real numbers
\newcommand{\myparagraph}[1]{\smallskip\noindent\textbf{#1}}

\newcommand{\mcb}{\color{blue}}

\newtheorem{theorem}{Theorem}%[section]
\newtheorem{lemma}{Lemma}%[section]
\newtheorem{proposition}{Proposition}%[section]
\newtheorem{definition}{Definition}%[section]
\usepackage{tikz}
\usetikzlibrary{arrows}

\def\iccvPaperID{4088} % *** Enter the ICCV Paper ID here

\ificcvfinal\fi
\begin{document}
	
	%%%%%%%%% TITLE
	\title{Affine Subspace Clustering under the Affinely Independent Subspace Model}
	\title{Affine Subspace Clustering: Theoretical Analysis and Experimental Evaluation}
	\title{Affine Subspace Clustering: Geometric Analysis and Experimental Evaluation}
	\title{Geometric Analysis and Experimental Evaluation on Affine Subspace Clustering}
	\title{Theoretical Analysis and Experimental Evaluation on Affine Subspace Clustering}
	\title{Theoretical Analysis and Experimental Evaluation of Affine Subspace Clustering}
	\title{A Geometric Analysis of Affine Subspace Clustering}
	\title{Theoretical Analysis of Affine Subspace Clustering}
	\title{Subspace Clustering: Linear, or Affine? A Theoretical Analysis}
	\title{A Theoretical Analysis of Subspace Clustering: Linear, or Affine?}
	\title{A Theoretical Analysis on Self-Expressive Model for Subspace Clustering: Linear, or Affine?}
	\title{A Geometric Analysis on Self-Expressive Model for Subspace Clustering: Linear, or Affine?}
	\title{A Geometric Analysis of Subspace Clustering: Linear, or Affine?}
	\title{A Geometric Analysis of Affine Subspace Clustering}
	\title{A Theoretical Analysis on Self-Expressive Model for Subspace Clustering: Linear, or Affine?}
	\title{Do we need affine constraint in affine subspace clustering?}
	\title{Do We Need Affine Constraint in Subspace Clustering?}
	\title{When Do We Need Affine Constraint in Subspace Clustering?}
	\title{Does Affine Subspace Clustering Need the Affine Constraint?}
	\title{Is the Affine Constraint Needed for Affine Subspace Clustering?}
	\title{Is an Affine Constraint Needed for Affine Subspace Clustering?}
	
	\author{First Author\\
		Institution1\\
		Institution1 address\\
		{\tt\small firstauthor@i1.org}
		% For a paper whose authors are all at the same institution,
		% omit the following lines up until the closing ``}''.
		% Additional authors and addresses can be added with ``\and'',
		% just like the second author.
		% To save space, use either the email address or home page, not both
		\and
		Second Author\\
		Institution2\\
		First line of institution2 address\\
		{\tt\small secondauthor@i2.org}
	}
	
	\author{Chong You$^1$ \quad Chun-Guang Li$^2$\quad Daniel P. Robinson$^3$\quad Ren\'e Vidal$^4$\\
		$^1$EECS, University of California, Berkeley, CA, USA\\
		$^2$SICE, Beijing University of Posts and Telecommunications, Beijing, China\\
		$^3$Applied Mathematics and Statistics, Johns Hopkins University, MD, USA\\
		$^4$Mathematical Institute for Data Science, Johns Hopkins University, MD, USA
	}
	
	\maketitle
	\ificcvfinal\thispagestyle{empty}\fi
	%\thispagestyle{empty}

	%%%%%%%%% ABSTRACT
	\begin{abstract}
		Subspace clustering methods based on expressing each data point as a linear combination of other data points have achieved great success in computer vision applications such as motion segmentation, face and digit clustering. In face clustering, the subspaces are linear and subspace clustering methods can be applied directly. In motion segmentation, the subspaces are affine and an additional affine constraint on the coefficients is often enforced. However, since affine subspaces can always be embedded into linear subspaces of one extra dimension, it is unclear if the affine constraint is really necessary.
		%Variants designed for affine subspaces enforce an additional affine constraint on the combination coefficients.
		%(i.e., shifted subspaces) enforce such expressions to be affine through the addition of an affine constraint.
		% affine spaces are shifted subspaces.
		% to account for affine subspaces, i.e., linear subspaces that do not pass the origin.
		%However, despite the prevalence of affine subspaces in real applications, the prior study and usage of these affine variants have been limited.
		This paper shows, both theoretically and empirically, that when the dimension of the ambient space is high relative to the sum of the dimensions of the affine subspaces, the affine constraint has a negligible effect on clustering performance.
		%This is demonstrated as an implication of our theoretical analysis and through  experiments on real datasets.
		Specifically, our analysis provides conditions that guarantee the correctness of affine subspace clustering methods both with and without the affine constraint, and shows that these conditions are satisfied for high-dimensional data.
		Underlying our analysis is the notion of affinely independent subspaces, which not only provides geometrically interpretable correctness conditions, but also clarifies the relationships between existing results for affine subspace clustering.
	\end{abstract}
	
	\section{Introduction}
	\label{sec:introduction}
	
	An important feature of modern data in computer vision is high-dimensionality. Images taken with mega-pixel cameras, for example, can be regarded as data points in a space of several million dimensions. Despite their high-dimensionality, data that correspond to the same group, such as %images of the same face,
	facial images of a subject,
	can usually be described by a few generating factors.  Such data is said to have an intrinsic dimension that is much smaller than the ambient space.
	When several such groups exist in the data, each one lying in a low-dimensional structure that is approximately linear, the data can be modeled as samples drawn from a union of linear subspaces.
	The problem of learning such a union of linear subspaces from unlabeled data is known as \emph{subspace clustering} \cite{Vidal:Springer16} and has drawn a lot of attention in areas such as computer vision~\cite{Costeira:IJCV98,Vidal:IJCV08,Rao:PAMI10}, system identification \cite{Bako-Vidal:HSCC08,Vidal:ACC04}, and bioinformatics~\cite{McWilliams:DMKD14}.
	
	In recent years, subspace clustering methods based on a self-expressiveness property of the data \cite{Elhamifar:CVPR09} have achieved great success.
	The self-expressiveness property states that each data point can be expressed as a linear combination of some other points from the data set. That is,
	\begin{equation}\label{eq:self-representation-constraint}
	%\begin{split}
	\x_j = \X \c_j \ \text{for each $j$},~~\text{or equivalently,}~~
	\X = \X\C,
	%\end{split}
	\end{equation}
	where $\X = [\x_1, \cdots, \x_N] \in \RR^{D \times N}$ is the data matrix and $\C = [\c_1, \cdots, \c_N] \in \RR^{N \times N}$ is the matrix of coefficients.
	A matrix $\C$ that satisfies the equations in \eqref{eq:self-representation-constraint} is usually not unique, but there always exists solutions whose entries satisfy $c_{ij} \ne 0$ only if $\x_i$ and $\x_j$ are from the same subspace. Such representations are called \emph{subspace-preserving}~\cite{Soltanolkotabi:AS12,You:ICML15,Vidal:Springer16}. A subspace-preserving $\C$ produces an affinity matrix $\W = |\C|+|\C^\transpose|$ with correct connections, \ie, $w_{ij}\neq 0$ only if $\x_i$ and $\x_j$ are from the same subspace. Spectral clustering \cite{vonLuxburg:StatComp2007} can then be applied to $\W$ to cluster the data $\X$.
	%. Spectral clustering uses $\W$ to obtain a clustering of the data $\X$.
	
	To find subspace-preserving solutions,
	many papers have proposed to solve the optimization problem
	\begin{equation}\label{eq:self-representation}
	\min_{\C} f(\C) \st \X = \X \C, ~~\C \in \Omega,
	\end{equation}
	where $f(\cdot)$ is a regularizer and $\Omega\subseteq\RR^{N\times N}$.
	%$ is a feasible set for $\C$.
	Existing methods make different choices for the regularizer $f(\cdot)$. For example, the sparse subspace clustering (SSC) method \cite{Elhamifar:CVPR09,Elhamifar:TPAMI13} uses $f(\C) = \|\C\|_1 := \sum_{i,j}|c_{ij}|$ to seek a sparse solution $\C$;
	the low-rank representation (LRR) \cite{Liu:ICML10,Liu:TPAMI13} and low-rank subspace clustering (LRSC) \cite{Favaro:CVPR11, Vidal:PRL14} methods use $f(\C) = \|\C\|_*$ to encourage $\C$ to be low-rank; and
	the least squares regression (LSR) \cite{Lu:ECCV12} and efficient dense subspace clustering (EDSC) \cite{Ji:WCACV14} methods use $f(\C) = \|\C\|_F^2$, as the optimization problem \eqref{eq:self-representation} with this regularization has a closed form solution.
	These methods have achieved excellent performance in many practical applications~\cite{Elhamifar:TPAMI13,You:CVPR16-EnSC,Li:TIP17,Ji:NIPS17,You:ECCV18,Zhang:ICML19,Li:CVPR19-subspace,Zhang:CVPR19}
	% Many works have provided theoretical justifications for the empirical success of these methods.
	and have accompanying theoretical support to justify their correctness in subspace detection~\cite{Soltanolkotabi:AS12, Lu:ECCV12, Wang:NIPS13-LRR+SSC, Liu:TPAMI13, Soltanolkotabi:AS14, Wang:ICML15, Wang:JMLR16, Yang:ECCV16, You:CVPR16-SSCOMP, You:CVPR16-EnSC, Tsakiris:ICML18, Tschannen:TIT18, Zhang:JMLR19,Robinson:arxiv19}. % detection subspace.
	In particular, it has been proven that all of these methods produce a subspace-preserving $\C$ when the subspaces are \emph{independent} (Definition \ref{def:independent-subspaces}) and $f(\cdot)$ satisfies the enforced block diagonal (EBD) conditions on $\Omega$ (Definition \ref{def:EBD}) \cite{Lu:ECCV12}.
	
	%Besides, there have also been works that use mixed regularizations to combine benefits from different regularizers.
	%For example, \cite{Wang:NIPS13-LRR+SSC} and \cite{You:CVPR16-EnSC} use $f(\C) = \lambda \|\C\|_1 + \|\C\|_*$ and $f(\C) = \lambda \|\C\|_1 + \frac{1-\lambda}{2}\|\C\|_F^2$, respectively, to balance the subspace-preserving property and the connectivity of the affinity graph.
	
	%We will review the concept of independent subspaces and EBD conditions in Section~\ref{sec:background}.

	\myparagraph{Affine subspace clustering.}
	Despite the great success of subspace clustering methods based on \eqref{eq:self-representation}, {the assumption that the subspaces are linear is often too restrictive because in many applications the subspaces do not pass through the origin, i.e., they are \emph{affine}.}
	%the union-of-subspace model is often too restrictive in practice because it assumes that subspaces pass through the origin of the coordinate system. In many applications, the subspaces do not pass through the origin, i.e., they are \emph{affine}.
	For example, in motion segmentation, feature point trajectories corresponding to the same rigid moving object lie approximately in a three dimensional affine subspace \cite{Tomasi:IJCV92}. Properly exploiting such affine structure is expected to boost clustering performance. Indeed, \cite{Elhamifar:TPAMI13,Ji:WCACV14,Jiang:COA18} address the motion segmentation problem by the following optimization problem in lieu of \eqref{eq:self-representation}:
	\begin{equation}\label{eq:affine-self-representation}
	\min_{\C} f(\C) \st \X = \X \C, ~\1^\transpose \C = \1^\transpose, ~\C \in \Omega.
	\end{equation}
	Here, $\1$ is a vector of length $N$ whose entries are all ones.
	The additional constraint $\1^\transpose \C = \1^\transpose$ imposes that the self-expressions use affine combinations rather than linear combinations, which is motivated from the observation that each point in an affine subspace can be expressed as an affine combination of other points in this affine subspace.

	{The effectiveness of methods based on \eqref{eq:affine-self-representation} demonstrated in \cite{Elhamifar:CVPR09,Elhamifar:TPAMI13,Ji:WCACV14} calls for the following theoretical question:
	}
	%The effectiveness of this method as demonstrated by \cite{Elhamifar:CVPR09,Elhamifar:TPAMI13,Ji:WCACV14} calls for the following theoretical question:
	%\begin{itemize}[topsep=0.3em]
	%	\item What conditions on the affine subspaces ensure that solutions to \eqref{eq:affine-self-representation} are subspace-preserving?
	%\end{itemize}
	\begin{quote}
		\em What conditions on the affine subspaces ensure that solutions to \eqref{eq:affine-self-representation} are subspace-preserving?
	\end{quote}
	{While this question has received a lot of attention in the case of linear subspaces, where one analyzes solutions to \eqref{eq:self-representation}, existing results for affine subspaces are surprisingly scarce.} For instance, \cite{Tsakiris:AffinePAMI17} provides an algebraic-geometric analysis of algebraic subspace clustering (ASC) \cite{Vidal:PAMI05} for affine subspaces, but the analysis does not extend to methods based on 
	\eqref{eq:affine-self-representation}. Then, while \cite{Elhamifar:CVPR09} provides a condition for SSC in terms of the homogeneous embedding of the affine subspaces, the condition does not provide a clear insight about the geometry of the original subspaces. Finally, while \cite{Li:JSTSP18} presents an analysis of SSC that has clear geometric interpretations, the analysis is restricted to SSC and it is unclear whether it is applicable to more general regularizers $f(\cdot)$ in \eqref{eq:affine-self-representation}.
	%does not apply to more general regularizers $f(\cdot)$.
	
	%In contrast to the abundance of results for linear subspace clustering, prior analysis for affine subspace clustering is surprisingly scarce. While a condition for SSC in terms of the homogeneous embedding of the affine subspaces is presented in \cite{Elhamifar:CVPR09}, this condition does not provide a clear insight about the geometry of the original subspaces.
	%Two recent works \cite{Tsakiris:AffinePAMI17,Li:JSTSP18} provide conditions that have clear insights in the original data space for the algebraic subspace clustering (ASC) \cite{Vidal:PAMI05} and SSC, respectively.
	%However, the analysis in \cite{Tsakiris:AffinePAMI17,Li:JSTSP18} is dedicated to ASC and SSC, respectively, and does not apply to \eqref{eq:affine-self-representation} for a general $f(\cdot)$.
	%%A recent work \cite{Li:JSTSP18} presents a geometric analysis for SSC that has clear interpretations.
	%%However, the analysis in \cite{Li:JSTSP18} is restricted to SSC and does not apply to a general regularizer $f(\cdot)$.

	\myparagraph{Is the affine constraint needed?}
	It is tempting to conclude that one should always use the formulation in \eqref{eq:affine-self-representation} rather than \eqref{eq:self-representation} in dealing with affine subspaces.
	Surprisingly, the majority of papers in the existing literature \cite{Lu:ECCV12,Liu:TPAMI13,Lu:ICCV13-TraceLasso,Soltanolkotabi:AS14,You:CVPR16-EnSC} adopt \eqref{eq:self-representation} in their experiments, even when datasets are affine. This calls for an explanation as to why the formulation in  \eqref{eq:self-representation} may work well for affine subspaces at all, and whether the affine constraint in \eqref{eq:affine-self-representation} is really needed.
	The former question may be answered by arguing that any $d$-dimensional affine subspace can be regarded as a subset of the $d+1$-dimensional linear subspace that contains the affine subspace, which justifies the application of linear subspace clustering methods to affine subspaces.
	Nonetheless, the following theoretical question has not been answered:
	%\begin{itemize}[topsep=0.3em]
	%	\item What conditions on the affine subspaces guarantee that solutions to \eqref{eq:self-representation} are subspace-preserving?
	%\end{itemize}
	\begin{quote}
		\em What conditions on the affine subspaces ensure that solutions to \eqref{eq:self-representation} are subspace-preserving?
	\end{quote}
	The answer to this question may help demystify the role of the affine constraint in \eqref{eq:affine-self-representation} and answer the question of when, and whether, it is needed for affine subspace clustering.

	\myparagraph{Contributions.}
	In this paper, we show that if the dimension of the ambient space is high enough, then both \eqref{eq:self-representation} and \eqref{eq:affine-self-representation} are guaranteed to produce subspace-preserving solutions under the model that the data points are drawn from a union of affine subspaces that are generated at random from the ambient space.
	This result justifies the usage of both \eqref{eq:self-representation} and \eqref{eq:affine-self-representation} for affine subspace clustering.
	It also suggests that the affine constraint in \eqref{eq:affine-self-representation} may not be needed when dealing with high-dimensional data, thus explaining the popularity of \eqref{eq:self-representation} in the existing literature.
	To verify that high-dimensionality plays a key role in drawing this conclusion, we conduct experiments on applications with both low-dimensional and high-dimensional ambient spaces, and show that the gap in performance between \eqref{eq:self-representation} and \eqref{eq:affine-self-representation} is usually prominent in the former case and often negligible in the latter case.
	
	Our discovery is important for practitioners seeking to choose an appropriate formulation for their problem.
	Solving the formulation with the affine constraint in \eqref{eq:affine-self-representation} is sometimes not as easy as solving the one without.
	For example, while an algorithm that can handle a million data points for SSC without the affine constraint has been developed in \cite{You:CVPR16-EnSC}, it cannot be easily adapted to handle the affine constraint for which existing solvers can only handle $\sim\!10,\!000$ data points \cite{Elhamifar:TPAMI13,Pourkamali:arXiv18}.
	%For example, while many efficient algorithms exist in the literature \cite{Efron:AS04, Lee:NIPS06, Figueiredo:STSP07, You:CVPR16-SSCOMP} %\cite{Pati:Asilomar93, You:CVPR16-EnSC}
	%for solving the optimization problem associated with SSC \cite{Elhamifar:CVPR09} without the affine constraint, none of them can be easily adapted to handling the affine constraint.\footnote{RV: I object. The ADMM for Basis pursuit can be easily extended.}
	Moreover, some of the methods, such as SSC-OMP \cite{You:CVPR16-SSCOMP} and $\ell_0$-SSC \cite{Yang:ECCV16}, cannot explicitly handle the affine constraint at all.
	For such methods, our result suggests that the affine constraint may not be needed at all and that the simpler model in \eqref{eq:self-representation} may be equally good.
	
	Our theoretical analysis is based on a novel approach to analyzing the affine subspace clustering problem that utilizes  the notion of \emph{affinely independent subspaces} developed in \cite{Li:JSTSP18}. This notion characterizes the arrangement of a collection of affine subspaces and has a clear geometric interpretation. Our results based on this notion provide geometric insights into the regimes where affine subspace clustering is easy for  self-expressiveness based methods.
	%Besides, our result makes it easier to understand the relationship between several existing conditions \cite{Elhamifar:CVPR09,Tsakiris:AffinePAMI17,Li:JSTSP18} for the correctness of affine subspace clustering.
	Besides, our analysis establishes several properties of affinely independent subspaces (\eg, %\ie, Theorem
	Lemma~\ref{theorem:independence-equivalency} and \ref{thm:relation-span-independence-affine-independence}),
	which makes it possible to compare several existing conditions \cite{Elhamifar:CVPR09,Tsakiris:AffinePAMI17,Li:JSTSP18} for the correctness of affine subspace clustering.

	\section{Background}
	\label{sec:background}

	This section provides the background for our theoretical analysis of formulations \eqref{eq:self-representation} and \eqref{eq:affine-self-representation} for affine subspace clustering, including a review of existing theoretical results for linear subspace clustering (Section \ref{sec:independent-subspace}) as well as the basics of affine geometry (Section \ref{sec:affine-independent-subspace}).

	\subsection{Subspace clustering under the independent subspace model}
	\label{sec:independent-subspace}
	
	A well-known result for linear subspace clustering is that the solution to \eqref{eq:self-representation} is subspace-preserving when the subspaces are independent and $f(\cdot)$ satisfies the Enforced Block Diagonal (EBD) conditions, as defined next.
	\begin{definition}[Independent linear subspaces \cite{Vidal:Springer16}]
		A collection of linear subspaces $\{\cS_{\ell}\}_{\ell=1}^n$ is said to be independent if $\dim(\spann(\cup_{\ell=1}^{n}\cS_{\ell})) = \sum_{\ell=1}^n\dim(\cS_{\ell}) $.
		\label{def:independent-subspaces}
	\end{definition}

	%A review of the definition of independent subspaces along with the basics of linear geometry is provided in the appendix. The EBD condition is defined as follows.
	
	%\begin{definition}[Enforced Block Diagonal (EBD) conditions \cite{Lu:ECCV12}]\label{def:EBD}
	%	Let $f(\cdot)$ be a function defined on a set $\Omega$ of square matrices.
	%	We say that $f(\cdot)$ satisfies the EBD conditions if for any $\C \in \Omega$, we have
	%	\begin{itemize}[topsep=0.1em,itemsep=0.1em,leftmargin=*]
	%		\item $\P^\transpose \C \P \in \Omega$ and $f(\C) = f(\P^\transpose \C \P)$ for any permutation matrix $\P$, and
	%		\item $\begin{bmatrix}
	%		\C_1 \!& \! \0 \\ \0 \!& \! \C_2
	%		\end{bmatrix} \in \Omega$ and $f\left(\begin{bmatrix}
	%		\C_1 \!& \!\C_3 \\ \C_4 \!& \!\C_2
	%		\end{bmatrix}\right) \ge f\left(\begin{bmatrix}
	%		\C_1 \!& \! \0 \\ \0 \!& \! \C_2
	%		\end{bmatrix}\right)$ for any partition of $\C$ so that $\C_1$ and $\C_2$ are square matrices, with equality holding if and only if $\C_3= \C_4= \0$.
	%	\end{itemize}
	%\end{definition}
	
	\begin{definition}[EBD conditions \cite{Lu:ECCV12}]
		\label{def:EBD}
		%\footnote{\color{red}I think the definition is not mathematically very precise. I would say: We say that $f(\cdot)$ satisfies the EBD condition if (1) The set $\Omega$ is closed under permutations and the function $f$ is permutation invariant, \ie, if $C\in\Omega$ then $P^\top C P\in\Omega$ and $f(C) = f(P^\top C P)$ for any permutation matrix $P$, and (2) the set $\Omega$ includes the set of all block-diagonal matrices and the function $f$ is block-diagonally dominant, \ie blah blah}
		%	Let $f(\cdot)$ be a function defined on a set $\Omega$ of square matrices.
		%	We say that $f(\cdot)$ satisfies the EBD conditions if for any $\C \in \Omega$, we have
		%	We say that a function $f(\cdot)$ defined on a set $\Omega$ of square matrices satisfies the EBD conditions if
		Let $\Omega \subset \RR^{N\times N}$. A function $f : \Omega \to \RR$ is said to satisfy the EBD conditions if
		\begin{itemize}[topsep=0.1em,itemsep=0.1em,leftmargin=*]
			\item $\Omega$ is closed under permutations and $f$ is permutation invariant, \ie, for any $\C \in \Omega$ we have $\P^\transpose \C \P \in \Omega$ and $f(\C) = f(\P^\transpose \C \P)$ for any permutation matrix $\P$, and
			\item for any partition $\C = \begin{bmatrix}
			\C_1 \!& \!\C_3 \\ \C_4 \!& \!\C_2
			\end{bmatrix}$ of any matrix $\C \in \Omega$ such that $\C_1$ and $\C_2$ are square matrices we have
			
			%\item $\Omega$ is closed under setting the off-diagonal blocks to zero and $f$ is block-diagonally dominant, \ie, if $\C \in \Omega$ then
			$\begin{bmatrix}
			\C_1 \!& \! \0 \\ \0 \!& \! \C_2
			\end{bmatrix} \in \Omega$
			and
			$f\left(\begin{bmatrix}
			\C_1 \!& \!\C_3 \\ \C_4 \!& \!\C_2
			\end{bmatrix}\right) \ge f\left(\begin{bmatrix}
			\C_1 \!& \! \0 \\ \0 \!& \! \C_2
			\end{bmatrix}\right)$
			
			%for any partition of $\C$ so that $\C_1$ and $\C_2$ are square matrices,
			with equality holding if and only if $\C_3= \C_4= \0$.
		\end{itemize}
	\end{definition}
	
	%The formal result for subspace clustering under the independent subspace model, {\mcb upon which our analysis for affine subspace clustering is based,} is presented next.
	{More formally, the main result for subspace clustering under the independent subspace model is the following.}\footnote{Several recent works \cite{Lu:TPAMI18,Xin:TSP18} show that the theorem holds for a broader range of $f$. All of our results hold with all such $f$ as well.
		%We note that a recent work \cite{Lu:TPAMI18} introduces a generalized version of the EBD conditions that cover a broader range of self-expressiveness based methods.
		%All of our results hold with that version of EBD as well.
	}
	\begin{theorem}[\cite{Lu:ECCV12}]\label{thm:subspace-clustering-EBD}
		Let $\X$ be the data matrix whose columns are drawn from a union of independent linear subspaces $\{\cS_{\ell}\}_{\ell=1}^n$.
		If $f$ satisfies the EBD conditions, then any solution to \eqref{eq:self-representation} is subspace-preserving.
	\end{theorem}

	%We note that a recent work \cite{Lu:TPAMI18} introduces a generalized version of the EBD conditions which cover a broader range of self-expressiveness based methods.
	%All our results hold with that version of EBD as well.

	\subsection{Affine geometry and affinely independent subspaces}
	\label{sec:affine-independent-subspace}
	%\subsection{Affine Geometry and Affine Independence}
	%\subsection{Affine Geometry}
	
	% We begin\footnote{\color{red}Should 2.1 and 2.2 be swapped, or should the word 'begin' be changed as we are no longer beginning the section/paper} with some basic concepts in affine geometry.
	
	%{\mcb We provide a review of the definition of independent subspaces along with the basics of linear geometry in the appendix, but review some basic concepts in affine geometry in this subsection.}
	
	Here, we review some basic concepts in affine geometry.
	\begin{itemize}[topsep=0.0em,itemsep=0.0em,leftmargin=*]
		%	\item \emph{Affine combination.} A point $\x \in \RR^D$ is an {affine combination} of points $\{\x_j \in \RR^D\}_{j=1}^k$ if $\x = \sum_{j=1}^k c_j \x_j$ and $\sum_{j=1}^k c_j = 1$.
		\item \emph{Affine subspace.} A nonempty set $\cA \subseteq \RR^D$ is an {affine subspace} if and only if every affine combination of points from $\cA$ lies in $\cA$.
		Equivalently, an affine subspace is a nonempty subset $\cA \subseteq \RR^D$ of the form $\cA = \x_0 + \cS := \{\x_0 + \x : \x \in \cS\}$, where $\cS \subseteq \RR^D$ is a linear subspace and $\x_0 \in \RR^D$ is a point.
		%In particular,
		%since the set $\cS$ above is independent of $\x_0$, we can
		The subspace $\cS$ associated with
		%an affine space
		$\cA$ is denoted by $\cT(\cA)$ and called the
		%; we call $\cT(\cA)$ the
		\textbf{direction subspace}.\footnote{Note that any linear subspace is also an affine subspace. In particular, given an affine subspace $\cA$, the following three statements are equivalent: (i) $\cA$ is a linear subspace; (ii) $\0 \in \cA$; and (iii) $\cT(\cA) = \cA$.} % of $\cA$.
	\end{itemize}
	%\noindent
	%Note that any linear subspace is also an affine subspace. In particular, given an affine subspace $\cA$, the following three statements are equivalent: (i) $\cA$ is a linear subspace; (ii) $\0 \in \cA$; and (iii) $\cT(\cA) = \cA$.
	
	%an affine subspace $\cA$ is a linear subspace: a) if and only if $\0 \in \cA$, and\footnote{RV: isn't it or instead of and Chong: ``and'' makes more sense?} b) if and only if $\cT(\cA) = \cA$.
	
	\begin{itemize}[topsep=0.0em,itemsep=0.0em,leftmargin=*]
		\item \emph{Affine hull (affine span). }The {affine hull/span} of a set $\cX \in \RR^D$, denoted as $\aff(\cX)$, is the intersection of all affine subspaces containing $\cX$. Equivalently, $\aff(\cX)$ is the set of all affine combinations of points in $\cX$.
		\item \emph{Affine independence.} A set of points $\{\x_j \in \RR^D\}_{j=1}^m$ is affinely independent if and only if $\sum_{j=1}^m c_j \x_j = \0$ and $\sum_{j=1}^m c_j = 0$ implies $c_j = 0$ for all $j \in \{1, \cdots, m\}$.
		\item \emph{Affine basis.} A set of points $\{\x_j \in \RR^D\}_{j=1}^m$ is an affine basis of $\cA$ if and only if it is affinely independent and its affine hull is $\cA$.
		\item \emph{Affine dimension.} {The dimension of an affine subspace $\cA$, denoted as $\dim(\cA)$, is defined as the dimension of its direction subspace, \ie, $\dim(\cA) \doteq \dim (\cT(\cA))$.}\footnote{Note that if $\dim(\cA) = m$, then any basis of $\cA$ has $m+1$ elements. Note also that this definition of dimension for affine subspaces generalizes the definition of dimension for linear subspaces. Specifically, any linear subspace can also be considered as an affine subspace, and its dimension as an affine subspace is equal to its dimension as a linear subspace.}\end{itemize}

	We now introduce the concepts of affine disjointness of two affine subspaces.%\footnote{RV: avoid affine twice. dpr: also, affine disjointness of two affine subspaces.}
	\begin{definition}[Affinely disjoint subspaces \cite{Li:JSTSP18}]
		Two affine subspaces $\cA$ and $\cA'$ are said to be affinely disjoint if and only if $\cA \cap \cA' = \emptyset$ and $\cT(\cA) \cap \cT(\cA') = \{\0\}$.
		\label{def:affinely-disjoint-affine-subspaces}
	\end{definition}
	Equivalently, two affine subspaces $\cA$ and $\cA'$ are affinely disjoint if and only if $\dim(\aff(\cA \cup \cA')) = \dim(\cA) + \dim(\cA')  + 1$ \cite{Li:JSTSP18}.
	For example, two $1$-dimensional affine subspaces in $\RR^3$ are affinely disjoint if and only if they are \emph{skewed} (\ie, neither parallel nor intersecting).
	\begin{definition}[Affinely independent subspaces \cite{Li:JSTSP18}]
		A collection of affine subspaces $\{\cA_{\ell}\}_{\ell=1}^n$ is said to be affinely independent if $\dim(\aff(\cup_{\ell=1}^{n}\cA_{\ell})) + 1 = \sum_{\ell=1}^n\dim(\cA_{\ell}) + n$.
		\label{def:affinely-independent-affine-subspaces}
	\end{definition}
	For an arbitrary collection of affine subspaces $\{\cA_{\ell}\}_{\ell=1}^n$, it has been shown in \cite{Li:JSTSP18} that
	\begin{equation}\label{eq:affine-dim-bound}
	\dim(\aff(\cup_{\ell=1}^{n}\cA_{\ell})) + 1 \le \sum_{\ell=1}^n\dim(\cA_{\ell}) + n.
	\end{equation}
	Therefore, the collection $\{\cA_{\ell}\}_{\ell=1}^n$ is affinely independent if and only if the affine subspaces are in an arrangement {such} that the dimension of the affine hull of their union is maximized.
	
	The concepts of affinely disjoint and affinely independent are closely related.
	Specifically, the set of affine subspaces $\{\cA_{\ell}\}_{\ell=1}^n$ is affinely independent if and only if for any  %disjoint nonempty subsets of $\{1, \cdots, n\}$, i.e., $\{\II', \II''\}
	subsets $\II',\II'' \subseteq \{1, \cdots, n\}$ with $\II' \cap \II'' = \emptyset$ it holds that the affine subspaces $\aff(\cup_{\kappa \in \II'}\cA_\kappa)$ and $\aff(\cup_{\kappa \in \II''}\cA_\kappa)$ are affinely disjoint.
	From this result, if the collection $\{\cA_{\ell}\}_{\ell=1}^n$ is affinely independent then they are pairwise affinely disjoint. The converse of this statement is not true.
	%In the literature, affine subspace clustering has not been received much attention. In the seminal work \cite{Elhamifar:CVPR09}, an affine constraint is added in SSC and the subspace-preserving guarantee based on the homogenously embedded data points has been given. However, the required geometric condition for the data points in the original space is unclear yet. In \cite{Tsakiris:AffinePAMI17}, theoretical guarantee for algebraic geometry based affine subspace clustering has been developed. However, the geometric condition in \cite{Tsakiris:AffinePAMI17} is not suitable for the self-expressiveness based subspace clustering approach in \eqref{eq:self-representation} and \eqref{eq:affine-self-representation}. Recently, a set of geometric conditions to guarantee the subspace-preserving solution for affine sparse subspace clustering have been established in \cite{Li:JSTSP18}, based on the affine independence assumption. However, it considers only sparsity based subspace clustering, rather than the general affine subspace clustering approach in \eqref{eq:self-representation} and \eqref{eq:affine-self-representation}. Moreover, the essential connection between the geometric condition in  \cite{Elhamifar:CVPR09} and the affine independence condition in \cite{Li:JSTSP18} is still unclear.
	
	\section{Affine Subspace Clustering Under the Affinely Independent Subspace Model}
	\label{sec:geometric_results}
	
	In this section, we establish geometric conditions under which the solutions to the optimization problems in \eqref{eq:self-representation} and in \eqref{eq:affine-self-representation} are subspace-preserving.
	%We then discuss the interpretation of these results and their implications for affine subspace clustering.
	The problem of \emph{affine subspace clustering} is formally defined as follows.
	\begin{definition}[Affine subspace clustering]
		Given a data matrix $\X = [\x_1, \cdots, \x_N] \in \RR^{D \times N}$ whose columns lie in a union of unknown affine subspaces $\{\cA_{\ell}\subseteq \RR^D\}_{\ell=1}^n$ of dimension $\{d_\ell < D\}_{\ell=1}^n$, affine subspace clustering is the problem of segmenting the data points into groups such that each group contains points from the same affine subspace.
		\label{def:affine-subspace-clustering}
	\end{definition}
	Since linear subspaces are a particular case of affine subspaces, the affine subspace clustering problem is a generalization of the linear subspace clustering problem.
	Throughout our theoretical analysis, we assume that the optimization problems \eqref{eq:self-representation} and \eqref{eq:affine-self-representation} are always feasible.
	This assumption does not impose stringent restrictions.
	For example, this is satisfied for arbitrary $f(\cdot)$ and $\X$ when $\Omega = \RR^{N \times N}$.
	%Throughout our theoretical analysis, we assume that {\mcb $\X$ contains at least $d_\ell+2$ points from $\cA_{\ell}$ for each $\ell$, including $d_\ell+1$ affinely independent points.}
	%In addition, we assume that the optimization problems \eqref{eq:self-representation} and \eqref{eq:affine-self-representation} are always feasible.
	%This assumption does not impose stringent restrictions.
	%For example, this is automatically satisfied for both \eqref{eq:self-representation} and \eqref{eq:affine-self-representation} when $\Omega = \RR^{N \times N}$.
	
	\subsection{Affine subspace clustering via formulation \eqref{eq:affine-self-representation}}
	
	We will show that the solutions to \eqref{eq:affine-self-representation} are subspace-preserving  under the affinely independent subspace model.
	Our analysis is based on the observation that applying \eqref{eq:affine-self-representation} to data $\X$ is equivalent to a two-step approach of first computing the homogeneous embedding of $\X$ as
	$\hbar(\X) := [\hbar(\x_1), \cdots, \hbar(\x_N)]$, where $\hbar: \RR^D \to \RR^{D+1}$ is the homogeneous embedding defined as
	\begin{equation}
	\hbar(\x) := [\x^\transpose, 1]^\transpose,
	\end{equation}
	and then solving the optimization problem in \eqref{eq:self-representation} but with $\X$ replaced by $\hbar(\X)$.
	Therefore, applying \eqref{eq:affine-self-representation} to data lying in affine subspaces $\{\cA_{\ell}\}_{\ell=1}^n$ is equivalent to applying \eqref{eq:self-representation} to data lying in embedded spaces $\{\hbar(\cA_{\ell})\}_{\ell=1}^n$, where $\hbar(\cA_{\ell}) := \{\hbar(\x): \x \in \cA_{\ell}\}$.
	The next result shows that each embedded space $\hbar(\cA_{\ell})$ is an affine subspace of dimension $d_\ell$ in $\RR^{D+1}$.
	Also, the linear subspace $\spann(\hbar(\cA_{\ell}))$ that contains $\hbar(\cA_{\ell})$ as a subset has dimension $d_\ell + 1$.
	\begin{lemma}\label{thm:dim-affine-vs-embed-span}
		Let $\cA$ be an arbitrary affine subspace in $\RR^D$. %We have that
		Then, a) $\hbar(\cA)$ is an affine subspace with $\dim(\hbar(\cA)) = \dim(\cA)$, and b) $\dim(\spann(\hbar(\cA)) = \dim(\cA) + 1$.
		%\begin{itemize}
		%\item $\hbar(\cA)$ is an affine subspace and $\dim(\hbar(\cA)) = \dim(\cA)$.
		%\item $\dim(\spann(\hbar(\cA)) = \dim(\cA) + 1$.
		%\end{itemize}
	\end{lemma}
	By this result, the embedded data $\hbar(\X)$ lies in a union of \emph{linear} subspaces $\{\spann(\hbar(\cA_{\ell}))\}_{\ell=1}^n$ whose dimensions are one more than the dimensions of the corresponding affine subspace.
	%From this result, each linear subspaces $\spann(\hbar(\cA_{\ell}))$ has dimension $d_\ell + 1$, therefore is a proper subspace of the ambient space $\RR^{D+1}$.
	This allows us to derive a correctness condition for \eqref{eq:affine-self-representation} by applying Theorem~\ref{thm:subspace-clustering-EBD} to this collection of linear subspaces.
	Specifically, we have the following result.
	%\footnote{We note that a similar correctness condition is provided in \cite{Elhamifar:CVPR09} for the case of $f(\C) = \|\C\|_1$.}.
	
	\begin{lemma}\label{thm:affine-subspace-clustering-on-affine-EBD}
		Let $\X$ be the data matrix in Definition~\ref{def:affine-subspace-clustering}.
		If $\{\spann(\hbar(\cA_{\ell}))\}_{\ell=1}^n$ is linearly independent and $f$ satisfies the EBD conditions, then any solution to \eqref{eq:affine-self-representation} is subspace-preserving.
	\end{lemma}
	
	We note that the same correctness condition is provided in \cite{Elhamifar:CVPR09} for the case of $f(\C) = \|\C\|_1$. However, the condition that $\{\spann(\hbar(\cA_{\ell}))\}_{\ell=1}^n$ is linearly independent is characterized by the span of the embedded affine subspaces in the homogeneously embedded ambient space, which makes it very difficult to interpret.
	We establish the following result, which shows that this condition is equivalent to the condition that the affine subspaces are affinely independent.
	\begin{lemma} % theorem
		\label{theorem:independence-equivalency}
		Let $\{\cA_{\ell}\}_{\ell=1}^n$ be a collection of affine subspaces.
		The subspaces $\{\spann(\hbar(\cA_{\ell}))\}_{\ell=1}^n$ are linearly independent if and only if $\{\cA_{\ell}\}_{\ell=1}^n$ are affinely independent.
	\end{lemma} % theorem

	By combining Lemma~\ref{thm:affine-subspace-clustering-on-affine-EBD} and %Theorem
	Lemma~\ref{theorem:independence-equivalency} we get the following result that gives  conditions under which the solution to \eqref{eq:affine-self-representation} is subspace-preserving.
	%\footnote{\color{red}Should't Theorem 3.1 be a Lemma, and Corollary 3.1 be a Theorem?}
	%\footnote{\color{blue}Li: I agree to replace Theorem 3.1 to Lemma 3.2, and change Corollary 3.1 be a Theorem 3.1, and the same for Theorem 3.2, Theorem 4.1, 4.2.}
	
	\begin{theorem} %corollary
		\label{thm:affine-subspace-clustering-on-affine-affinely-independent}
		Let $\X$ be the data matrix in Definition~\ref{def:affine-subspace-clustering}.
		If $\{\cA_{\ell}\}_{\ell=1}^n$ is affinely independent and $f$ satisfies the EBD conditions, then any solution to \eqref{eq:affine-self-representation} is subspace-preserving.
	\end{theorem} %corollary
	
	The condition that $\{\cA_{\ell}\}_{\ell=1}^n$ is affinely independent has a clearer geometric interpretation since it is defined directly in the original data space rather than in the homogeneously embedded space. Moreover, it is verifiable when the arrangement of affine subspaces is known, which helps us understand when \eqref{eq:affine-self-representation} is applicable. Besides, it also allows us to compare it with subspace-preserving conditions for the formulation in \eqref{eq:self-representation} as we will see in the next subsection.
	
	Finally, we point out that for a collection of affine subspaces to be affinely independent, the ambient dimension needs to be large enough  relative to the individual subspace dimensions and the number of subspaces. This is formally stated as our next result.
	\begin{proposition}
		\label{thm:condition-2-dimension-bound}
		Let $\{\cA_{\ell} \subseteq \RR^D\}_{\ell=1}^n$ be a collection of affine subspaces.
		If $\{\cA_{\ell}\}_{\ell=1}^n$ is affinely independent, then
		\vspace{-2mm}
		\begin{equation}\label{eq:condition-2-dimension-bound}
		D \ge \dim(\spann(\cup_{\ell = 1}^n \cA_{\ell})) \ge \sum_{\ell=1}^n \dim(\cA_{\ell}) + n - 1.
		\end{equation}
	\end{proposition}

	\subsection{Affine subspace clustering via formulation \eqref{eq:self-representation}}
	
	It may not be surprising that the formulation in \eqref{eq:self-representation}, although designed for linear subspaces, may also work for affine subspaces, since each affine subspace $\cA_{\ell}$ can be regarded as a subset of the linear subspace $\spann(\cA_{\ell})$.
	In other words, clustering data in the affine subspaces $\{\cA_{\ell}\}_{\ell=1}^n$ can be regarded as clustering data in the linear subspaces $\{\spann(\cA_{\ell})\}_{\ell=1}^n$.
	%The following result shows that the dimension of the linear subspace $\spann(\cA_{\ell})$ is related to the dimension of the affine subspace $\cA_{\ell}$ for each $\ell = 1, \cdots, n$.
	%\begin{lemma}\label{thm:dim-affine-vs-span}
	%	Let $\cA$ be an arbitrary affine subspace. If $\0 \in \cA$ (i.e., if $\cA$ is a linear subspace), then $\dim(\spann(\cA)) = \dim(\cA)$; otherwise, $\dim(\spann(\cA)) = \dim(\cA) + 1$.
	%\end{lemma}
	It is easy to see that the dimension of the linear subspace $\spann(\cA_{\ell})$ is related to the dimension of the affine subspace $\cA_{\ell}$ for each $\ell = 1, \cdots, n$. Specifically, if $\0 \in \cA$ (i.e., if $\cA$ is a linear subspace), then $\dim(\spann(\cA)) = \dim(\cA)$; otherwise, $\dim(\spann(\cA)) = \dim(\cA) + 1$.
	From this result, each linear subspace $\spann(\cA_{\ell})$ has dimension either $d_\ell$ (when $\0 \in \cA_{\ell}$) or $d_\ell + 1$ (when $\0 \notin \cA_{\ell}$).
	Note that if $D = d_\ell + 1$ and $\0 \notin \cA_{\ell}$ for some $\ell$, then the linear subspace $\spann(\cA_{\ell})$ becomes the ambient space $\RR^D$.
	In such cases, the problem of clustering data drawn from the linear subspaces $\{\spann(\cA_{\ell})\}_{\ell=1}^n$ is ill-posed.
	In all other cases, the subspaces $\{\spann(\cA_{\ell})\}_{\ell=1}^n$ are proper linear subspaces of $\RR^D$.
	By applying Theorem~\ref{thm:subspace-clustering-EBD} to this collection of linear subspaces, we get the following result.
	%which states that the solution to \eqref{eq:self-representation} is subspace-preserving if $\{\spann(\cA_{\ell})\}_{\ell=1}^n$ is linearly independent %(i.e., Condition~\ref{cnd:span-independent} holds)
	% and $f$ satisfies the EBD conditions. Formally, we have the following lemma.
	
	\begin{lemma}\label{thm:subspace-clustering-on-affine-EBD}
		Let $\X$ be the data matrix in Definition~\ref{def:affine-subspace-clustering}. If $\{\spann(\cA_{\ell})\}_{\ell=1}^n$ is linearly independent and $f$ satisfies the EBD conditions, then any solution to \eqref{eq:self-representation} is subspace-preserving.
	\end{lemma}
	Although Lemma~\ref{thm:subspace-clustering-on-affine-EBD}  establishes the correctness of \eqref{eq:self-representation} for affine subspace clustering, the condition that $\{\spann(\cA_{\ell})\}_{\ell=1}^n$ is linearly independent is not particularly insightful in terms of the geometry of the affine subspaces.
	The next result shows that under the affinely independent subspace model, the condition in Lemma~\ref{thm:subspace-clustering-on-affine-EBD} can be expressed in a form that has a clearer geometric interpretation.
	
	\begin{lemma} % theorem
		\label{thm:relation-span-independence-affine-independence}
		Let $\{\cA_{\ell}\}_{\ell=1}^n$ be a collection of affine subspaces such that $\0 \notin \cup_{\ell=1}^n \cA_{\ell}$.
		Then, the collection of linear subspaces $\{\spann(\cA_{\ell})\}_{\ell=1}^n$ is linearly independent if and only if the following two conditions hold:
		\begin{itemize}[topsep=0.1em,itemsep=0.1em]
			\item $\{\cA_{\ell}\}_{\ell=1}^n$ is affinely independent; and
			\item $\0 \notin \aff(\cup_{\ell=1}^n \cA_{\ell})$.
		\end{itemize}
	\end{lemma} % theorem
	
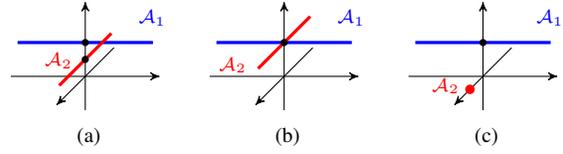
\begin{figure}[ht]
		%	\vskip 0.2in
		\begin{center}
			\subfigure[\label{fig:span-independent-equivalent-1}]{
				\begin{tikzpicture}[
				scale=0.9,
				axis/.style={very thick, ->, >=stealth'}
				]
				% axis
				\draw[axis,thin] (xyz cs:x=-1.1) -- (xyz cs:x=1.1);
				\draw[axis,thin] (xyz cs:y=-0.5) -- (xyz cs:y=1.1);
				\draw[axis,thin] (xyz cs:z=-1.1) -- (xyz cs:z=1.1);
				\draw[very thick, blue] (xyz cs:y=0.5,x=-1) -- (xyz cs:y=0.5,x=1) node[above] {$^{\cA_1}$};
				\fill[black] (xyz cs:y=0.5) circle (1.5pt);
				\draw[very thick, red] (xyz cs:y=0.25,z=-1) -- (xyz cs:y=0.25,z=1) node[above] {$^{\cA_2}$};
				\fill[black] (xyz cs:y=0.25) circle (1.5pt);
				\end{tikzpicture}
			}
			~
			\subfigure[\label{fig:span-independent-equivalent-2}]{
				\begin{tikzpicture}[
				scale=0.9,
				axis/.style={very thick, ->, >=stealth'}
				]
				% axis
				\draw[axis,thin] (xyz cs:x=-1.1) -- (xyz cs:x=1.1);
				\draw[axis,thin] (xyz cs:y=-0.5) -- (xyz cs:y=1.1);
				\draw[axis,thin] (xyz cs:z=-1.1) -- (xyz cs:z=1.1);
				\draw[very thick, blue] (xyz cs:y=0.5,x=-1) -- (xyz cs:y=0.5,x=1) node[above] {$^{\cA_1}$};
				\draw[very thick, red] (xyz cs:y=0.5,z=-1) -- (xyz cs:y=0.5,z=1) node[left] {$^{\cA_2}$};
				\fill[black] (xyz cs:y=0.5) circle (1.5pt);
				\end{tikzpicture}
			}
			~
			\subfigure[\label{fig:span-independent-equivalent-3}]{
				\begin{tikzpicture}[
				scale=0.9,
				axis/.style={very thick, ->, >=stealth'}
				]
				% axis
				\draw[axis,thin] (xyz cs:x=-1.1) -- (xyz cs:x=1.1);
				\draw[axis,thin] (xyz cs:y=-0.5) -- (xyz cs:y=1.1);
				\draw[axis,thin] (xyz cs:z=-1.1) -- (xyz cs:z=1.1);
				\draw[very thick, blue] (xyz cs:y=0.5,x=-1) -- (xyz cs:y=0.5,x=1) node[above] {$^{\cA_1}$};
				\fill[red] (xyz cs:z=0.5) circle (2pt)
				node[left] {$^{\cA_2}$};
				\fill[black] (xyz cs:y=0.5) circle (1.5pt);
				\end{tikzpicture}
			}
			\caption{Illustration of the conditions in Theorem~\ref{thm:relation-span-independence-affine-independence}.}
			\label{fig:span-independent-equivalent}
		\end{center}
		\vskip -0.2in
	\end{figure}

	The conditions that $\{\cA_{\ell}\}_{\ell=1}^n$ is affinely independent and that $\0 \notin \aff(\cup_{\ell=1}^n \cA_{\ell})$ are both needed in %Theorem
	Lemma~\ref{thm:relation-span-independence-affine-independence}.
	In Figure~\ref{fig:span-independent-equivalent-1} we show an example of two $1$-dimensional affine subspaces $\cA_1, \cA_2$ in $\RR^3$ for which the first condition is satisfied (i.e., $\{\cA_\ell\}_{\ell=1}^2$ is affinely independent), but the second condition is violated (i.e., $\0 \in \aff(\cup_{\ell=1}^2 \cA_{\ell})$).
	In Figure~\ref{fig:span-independent-equivalent-2} we show an example where the first condition is violated and the second condition is satisfied.
	In both of these examples we can easily see that $\spann(\cA_1)$ and $\spann(\cA_2)$ are not independent subspaces.
	In fact, it is impossible to find two $1$-dimensional affine subspaces in $\RR^3$ that satisfy both conditions in %Theorem
	Lemma~\ref{thm:relation-span-independence-affine-independence}.
	More generally, the following result shows that the ambient dimension needs to be sufficiently large in order for both conditions to hold.
	\begin{proposition}
		\label{thm:condition-1-dimension-bound}
		Let $\{\cA_{\ell} \subseteq \RR^D\}_{\ell=1}^n$ be a collection of affine subspaces.
		If $\{\cA_{\ell}\}_{\ell=1}^n$ is affinely independent and $\0 \notin \aff(\cup_{\ell=1}^n \cA_{\ell})$, then
		\begin{equation}\label{eq:condition-1-dimension-bound}
		D \ge \dim(\spann(\cup_{\ell = 1}^n \cA_{\ell})) = \sum_{\ell=1}^n \dim(\cA_{\ell}) + n.
		\end{equation}
	\end{proposition}
	%We provide the proof of Proposition~\ref{thm:condition-1-dimension-bound} in the appendix.
	{Finally, Figure~\ref{fig:span-independent-equivalent-3} gives an example of a $1$-dimensional subspace $\cA_1$ and a $0$-dimensional subspace $\cA_2$ for which} the conditions that $\{\cA_{\ell}\}_{\ell=1}^2$ is affinely independent and that $\0 \notin \aff(\cup_{\ell=1}^2 \cA_{\ell})$ are both satisfied.
	Note that in this particular example, the inequality in \eqref{eq:condition-1-dimension-bound} holds with equality.

	We now combine Lemma~\ref{thm:subspace-clustering-on-affine-EBD} and %Theorem
	Lemma~\ref{thm:relation-span-independence-affine-independence} to get the following important result, which gives conditions under which the solution to \eqref{eq:self-representation} is subspace-preserving.
	%\footnote{\color{red}Same comment as before, replacing Theorem 3.2 by a Lemma, and Corollary 3.2 by a theorem.}
	
	\begin{theorem} % corollary
		Let $\X$ be the data matrix in Definition~\ref{def:affine-subspace-clustering}.
		If $\{\cA_{\ell}\}_{\ell=1}^n$ is affinely independent, $\0 \notin \aff(\cup_{\ell=1}^n \cA_{\ell})$ and $f$ satisfies the EBD conditions, then any solution to \eqref{eq:self-representation} is subspace-preserving.
		\label{thm:subspace-clustering-on-affine-affinely-independent}
	\end{theorem} % corollary
	From this result, we can easily compare the subspace-preserving conditions for \eqref{eq:affine-self-representation} and \eqref{eq:self-representation}, as discussed next.

	\subsection{Comparison and discussion}

	%Corollary
	Theorem~\ref{thm:affine-subspace-clustering-on-affine-affinely-independent} and %Corollary
	Theorem~\ref{thm:subspace-clustering-on-affine-affinely-independent} establish conditions under which the solutions to \eqref{eq:affine-self-representation} and \eqref{eq:self-representation}, respectively, are subspace-preserving for affine subspace clustering.
	To compare the conditions in these two results, we see that both of them require the affine subspaces to be affinely independent, while %Corollary
	Theorem~\ref{thm:subspace-clustering-on-affine-affinely-independent} imposes an additional requirement that $\0 \notin \aff(\cup_{\ell=1}^n \cA_{\ell})$.
	As illustrated in Figure~\ref{fig:span-independent-equivalent}, there exist cases where the affine subspaces are affinely independent but the condition $\0 \notin \aff(\cup_{\ell=1}^n \cA_{\ell})$ is not satisfied.
	Therefore, the theoretical guarantees for \eqref{eq:affine-self-representation} apply to a broader range of problems than those for \eqref{eq:self-representation}.
	%This suggests that \eqref{eq:affine-self-representation} may be a more preferable method for addressing the affine subspace clustering problem.

	The difference between the conditions in %Corollary
	Theorem~\ref{thm:affine-subspace-clustering-on-affine-affinely-independent} and %Corollary
	Theorem~\ref{thm:subspace-clustering-on-affine-affinely-independent} can also be seen in terms of the regime where they can be satisfied.
	Specifically, Proposition~\ref{thm:condition-2-dimension-bound} shows that the conditions in %Corollary
	Theorem~\ref{thm:affine-subspace-clustering-on-affine-affinely-independent} may be satisfied only if
	%\begin{equation}\label{eq:dimension-condition-2-rewritten}
	$D \ge \sum_{\ell=1}^n d_\ell + n - 1$,
	%\end{equation}
	while Proposition~\ref{thm:condition-1-dimension-bound} shows that the conditions in %Corollary
	Theorem~\ref{thm:subspace-clustering-on-affine-affinely-independent} may be satisfied only if
	%\begin{equation}\label{eq:dimension-condition-1-rewritten}
	$D \ge \sum_{\ell=1}^n d_\ell + n$.
	%\end{equation}
	This again suggests that subspace-preserving recovery by \eqref{eq:affine-self-representation} may be easier than by \eqref{eq:self-representation}.
	This conclusion also aligns with our intuition: \eqref{eq:affine-self-representation} should work better than \eqref{eq:self-representation} as it is explicitly modeling the affine structure by means of the affine constraint in its formulation.
	
	So far, we have understood that the formulation in \eqref{eq:affine-self-representation} is advantageous over the formulation in \eqref{eq:self-representation}.
	However, for practical applications we would like to understand how significant this advantage is.
	In the next section, we show that both \eqref{eq:affine-self-representation} and \eqref{eq:self-representation} produce subspace-preserving solutions when the dimension of the ambient space is high enough relative to subspace dimensions, suggesting that the difference in their performance for such data may be negligible. %minimal.

	%\subsection{Why affinely independent? A closer look}
	%
	%So far, our analysis of \eqref{eq:self-representation} and \eqref{eq:affine-self-representation} for affine subspace clustering above takes the approach of applying existing analysis for linear subspace clustering, and then establishing the connection of the correctness conditions with the affinely independent condition.
	%
	%(Can we justify why affinely independent is a ``natural'' choice for the analysis of affine subspace clustering? Maybe by presenting the proof of Theorem~\ref{thm:affine-subspace-clustering-on-affine-EBD}? Maybe by taking closer look at the geometric interpretation of the model?)

	\section{Affine Subspace Clustering Under a Random Affine Subspace Model}
	\label{sec:random_results}
	
	In this section, we study the conditions under which the solution to \eqref{eq:self-representation} and \eqref{eq:affine-self-representation} are subspace-preserving when the affine subspaces are generated according to the following random affine subspace model.
	%\begin{definition}[Random Affine Subspace Model]
	%	Given an ambient dimension $D$ and subspace dimensions $\{d_\ell\}_{\ell=1}^n$, our random affine subspace model is defined as generating affine subspaces $\{\cA_{\ell}\}_{\ell=1}^n$, where $\cA_{\ell} = \w_{0,\ell}+\spann\{\w_{1,\ell}, \cdots, \w_{d_\ell,\ell}\}$ for each $\ell$ and the vectors $\{\w_{0,\ell}, \w_{1,\ell}, \cdots, \w_{d_\ell,\ell}\}_{\ell=1}^n$ are drawn independently and uniformly at random from the unit sphere of $\RR^D$.
	%	\label{def:random-affine-subspace-model}
	%\end{definition}
	%
	{
		\begin{definition}[Random Affine Subspace Model]
			A collection $\{\cA_{\ell}\}_{\ell=1}^n$ of $n$ affine subspaces of $\RR^D$ of dimensions $\{d_\ell < D\}_{\ell=1}^n$ is said to be drawn from the random affine subspace model if
			$\cA_{\ell} = \w_{0,\ell}+\spann\{\w_{1,\ell}, \cdots, \w_{d_\ell,\ell}\}$, where $\{\w_{0,\ell}, \w_{1,\ell}, \cdots, \w_{d_\ell,\ell}\}_{\ell=1}^n$ are drawn independently and uniformly at random from the unit sphere of $\RR^D$.
			\label{def:random-affine-subspace-model}
	\end{definition}}
	\vspace{-1mm}
	
	The random affine subspace model given by Definition~\ref{def:random-affine-subspace-model} is equivalent to drawing $n$ linear subspaces $\{\cS_{\ell}\}_{\ell=1}^n = \{\spann\{\w_{1,\ell},\dots,\w_{d_\ell,\ell}\}\}_{\ell=1}^n$ (i.e., the direction subspaces for the affine subspaces $\{\cA_{\ell}\}_{\ell=1}^n$) independently and uniformly at random from the ambient space of $\RR^D$ and then adding to each subspace $\cS_{\ell}$ a random vector $\w_{0,\ell}$ that is drawn independently and uniformly at random from the unit sphere of $\RR^D$.
	Each subspace $\cS_{\ell}$ has dimension $d_\ell$ with probability $1$, as does the corresponding affine subspace $\cA_{\ell}$ since $\dim(\cA_{\ell}) = \dim(\cS_{\ell})$.
	As long as $D > d_\ell$, $\cA_{\ell}$ is an affine (not linear) subspace with probability $1$, which can be seen from the fact that $\w_{0,\ell}$ is drawn independently from the subspace $\spann\{\w_{1,\ell}, \cdots, \w_{d_\ell,\ell}\}$.
	
	Recall from Proposition~\ref{thm:condition-2-dimension-bound} and Proposition~\ref{thm:condition-1-dimension-bound} that the dimension of the ambient space needs to be large enough in order for the geometric conditions in %Corollary
	Theorem~\ref{thm:affine-subspace-clustering-on-affine-affinely-independent} and %Corollary
	Theorem~\ref{thm:subspace-clustering-on-affine-affinely-independent} to be satisfied.
	The following two results show that such conditions are not only necessary but also sufficient under the random affine subspace model.

	\begin{lemma} % theorem
		If $D \ge \sum_{\ell=1}^n d_\ell+n - 1$, then the collection of affine subspaces $\{\cA_{\ell} \subseteq \RR^D\}_{\ell=1}^n$ drawn according to the random affine subspace model in Definition~\ref{def:random-affine-subspace-model} is affinely independent with probability $1$.
		\label{thm:affine-subspace-clustering-under-random}
	\end{lemma} % theorem
	
	\begin{lemma}
		If $D \ge \sum_{\ell=1}^n d_\ell+n$, then the collection of affine subspaces $\{\cA_{\ell} \subseteq \RR^D\}_{\ell=1}^n$ drawn according to the random affine subspace model in
		Definition~\ref{def:random-affine-subspace-model}
		is affinely independent with $\0 \notin \aff(\cup_{\ell=1}^n \cA_{\ell})$ with probability $1$.
		\label{thm:subspace-clustering-under-random}
	\end{lemma}

	By combining %Theorem
	Lemma~\ref{thm:affine-subspace-clustering-under-random} and Lemma %Theorem
	\ref{thm:subspace-clustering-under-random} with %Corollary
	Theorem~\ref{thm:affine-subspace-clustering-on-affine-affinely-independent} and %Corollary
	Theorem~\ref{thm:subspace-clustering-on-affine-affinely-independent} we get the following result.%\footnote{RV: Some condition on the data must be given: like sufficiently many points in generic position inside each subspace. Chong: the same comment applies to Sec 3. I added a paragraph at the beginning of Sec 3}
	
	\begin{theorem}\label{thm:affine-subspace-clustering-under-random-corollary}
		{
			Let $\{\cA_{\ell}\}_{\ell=1}^n$ be a collection of $n$ affine subspaces of $\RR^D$ of dimensions $\{d_\ell < D\}_{\ell=1}^n$ drawn}
		%Given an ambient dimension $D$ and subspace dimensions $\{d_\ell\}_{\ell=1}^n$, draw a collection of affine subspaces $\{\cA_{\ell}\}_{\ell=1}^n$ 
		according to the random affine subspace model in Definition~\ref{def:random-affine-subspace-model}.
		%Let $\X$ be a data matrix whose columns are drawn from
		%\footnote{\color{red}How are they drawn? Is this true for any distribution. Cannot be fixed because subspaces are not fixed} 
		Let $\X$ be an arbitrary data matrix whose columns lie in $\cup_{\ell = 1}^n \cA_{\ell}$.
		Assume that $f$ satisfies the EBD conditions.
		\begin{enumerate}[topsep=0.2em,itemsep=0.0em,label=(\roman*)]
			\item \label{itm:with-affine}If $D \ge \sum_{\ell=1}^n d_\ell+n - 1$, then any solution to \eqref{eq:affine-self-representation} is subspace-preserving with probability $1$.
			\item \label{itm:no-affine}If $D \ge \sum_{\ell=1}^n d_\ell+n$, then any solution to \eqref{eq:self-representation} is subspace-preserving with probability~$1$.
		\end{enumerate}
	\end{theorem}
	
	%Similarly, by combining Theorem~\ref{thm:subspace-clustering-under-random} and Corollary~\ref{thm:subspace-clustering-on-affine-affinely-independent}  we get the following result.
	
	%\begin{corollary}\label{thm:subspace-clustering-under-random-corollary}
	%	In the same setting as Corollary~\ref{thm:affine-subspace-clustering-under-random-corollary},
	%	if $f$ satisfies the EBD conditions and $D \ge \sum_{\ell=1}^n d_\ell+n$, then any solution to \eqref{eq:self-representation} is subspace-preserving with probability~$1$.
	%\end{corollary}
	
	Theorem~\ref{thm:affine-subspace-clustering-under-random-corollary} justifies our claim in the introduction that both \eqref{eq:affine-self-representation} and \eqref{eq:self-representation} produce subspace-preserving solutions when the dimension of the ambient space is large enough.
	In our experiments on synthetic data, we will show that the thresholds on the dimension $D$ as stated in Theorem~\ref{thm:affine-subspace-clustering-under-random-corollary} are tight for the case where $f(\cdot)=\|\cdot\|_F^2$ (\ie the LSR method). That is, the solution to LSR with and without the affine constraint is observed to be not subspace-preserving when the ambient dimension $D$ is smaller than $\sum_{\ell=1}^n d_\ell+n - 1$ and $\sum_{\ell=1}^n d_\ell+n$, respectively.
	This suggests that high-dimensionality of the ambient space is necessary for the solution to \eqref{eq:affine-self-representation} and \eqref{eq:self-representation} to be subspace-preserving in general.
	
	For real data, the affine subspaces usually do not satisfy the random affine subspace model, therefore the solution may not necessarily be subspace-preserving even if the ambient dimension is high enough.
	Nonetheless, we still observe that the difference in performance between \eqref{eq:affine-self-representation} and \eqref{eq:self-representation} is often small or negligible for high-dimensional data.
	We present such an investigation %an study
	in the next section.
	
	\section{Experiments}
	
	{\mcb % The goal is to compare (2) and (4) with $f$ set to be SSC, LRR and LSR.
		
		%Synthetic experiments to demonstrate the performance of the two methods. If we randomly generate $n$ subspaces each of dimension $d$ in ambient dimension $D$, translate each of them by a random vector, then we get a union of affinely independent affine subspaces if $D + 1 = n \cdot (d + 1)$ with probability $1$ (my conjecture). We will see that (4) gives subspace-preserving solution, but (2) will not (my conjecture). This verifies that (4) is better.
		
		% Real experiments to demonstrate that (4) is better than (2) on real data (hopefully)
	}
	
	We conduct experiments on synthetic datasets to verify our theoretical results and to further understand the behavior of formulations \eqref{eq:affine-self-representation} and \eqref{eq:self-representation} for affine subspace clustering.
	We also perform experiments on real datasets that include both low-dimensional and high-dimensional settings to better understand the difference in their performances.
	% between these two methods.
	
	The formulations \eqref{eq:affine-self-representation} and \eqref{eq:self-representation} encompass a wide range of methods that have been studied in the existing literature.
	For the purpose of this study, we restrict our attention to the SSC method (i.e., $f(\cdot) = \|\cdot\|_1$ and $\Omega=\{\C\in \RR^{N \times N}: \diag(\C)=\0\}$) and LSR method (i.e., $f(\cdot) = \tfrac{1}{2}\|\cdot\|_F^2$  and $\Omega=\RR^{N \times N}$).
	For both of them, the function $f(\cdot)$ satisfies the EBD conditions.
	To distinguish between \eqref{eq:affine-self-representation} and \eqref{eq:self-representation}, we will refer to the methods corresponding to \eqref{eq:affine-self-representation} as A-SSC and A-LSR, and the methods corresponding to \eqref{eq:self-representation} as SSC and LSR.
	In our experiments on synthetic data, we use CVX \cite{cvx} to solve the optimization problems associated with A-SSC and SSC, and use the closed form solutions given by $\begin{bmatrix} \X \\ \1^\transpose\end{bmatrix}^\dagger \begin{bmatrix} \X \\ \1^\transpose\end{bmatrix}$ and $\X^\dagger \X$, respectively, for A-LSR and LSR, where $\X^\dagger$ denotes the pseudoinverse of $\X$.
	
	For experiments on real data, we penalize the self-expressive residual instead of imposing the equality constraint to account for noise in the data.
	That is, for a chosen parameter $\lambda > 0$, instead of \eqref{eq:affine-self-representation} we use
	\begin{equation}\label{eq:affine-self-representation-noisy}
	\min_{\C} f(\C) + \tfrac{\lambda}{2} \|\X - \X \C\|_F^2 \!\!\!\st\!\! \1^\transpose \C = \1^\transpose, ~\C \in \Omega,
	\end{equation}
	and instead of \eqref{eq:self-representation} we use
	\begin{equation}\label{eq:self-representation-noisy}
	\min_{\C} f(\C) + \tfrac{\lambda}{2} \|\X - \X \C\|_F^2 \st ~\C \in \Omega.
	\end{equation}
	%In \eqref{eq:affine-self-representation-noisy} and \eqref{eq:self-representation-noisy}, $\lambda > 0$ is a parameter.
	For A-SSC and SSC, we follow \cite{Elhamifar:TPAMI13} and set $\lambda=\alpha/\mu_z$ where $\alpha$ is a parameter and $\mu_z$ is defined in \cite{Elhamifar:TPAMI13}, and solve the associated optimization problems via the alternating direction method of multipliers (ADMM) algorithm.
	For A-LSR and LSR the optimization problems have closed form solutions that can be found in \cite{Lu:ECCV12} for LSR and given for A-LSR by
	%\begin{equation}
	$\C = \lambda \W \X^\transpose \X + (\1^\transpose \v)^{-1}\v \v^\transpose$, %\frac{\v \v^\transpose}{\1^\transpose \v},
	%\end{equation}
	where $\W = (\lambda \X^\transpose \X + \I )^{-1}$ and $\v =\W \cdot\1$.

	\subsection{Experiments on synthetic data}
	
	We verify %Corollary
	Theorem~\ref{thm:affine-subspace-clustering-under-random-corollary} by generating affine subspaces according to the random model in Definition~\ref{def:random-affine-subspace-model} and sampling data points on the affine subspaces at random.
	Specifically, we first sample $n$ linear subspaces of dimension $d$ independently and uniformly at random from $\RR^D$.
	Then, we sample $N / n$ data points from the unit sphere of each subspace independently and uniformly at random.
	Finally, we generate $n$ vectors on the unit sphere of the ambient space, and add each one of them to all data points in the corresponding subspaces.
	This gives $N$ points lying in a union of randomly generated affine subspaces.
	In our experiments, we fix $d = 4$, $n = 5$, and $N = 100$ and vary the ambient dimension $D$ in $\{5, 6, \cdots, 30\}$.%the range $[5, 30]$.
	
	To evaluate the degree to which the subspace-preserving property is satisfied, we use the subspace-preserving rate (SPR) defined as
	$\frac{1}{N}\sum_{j=1}^N (\sum_{i=1}^N(w_{ij}\cdot |c_{ij}|) / \|\c_j\|_1)$,
	%\begin{equation}
	%\text{SPR} = \frac{1}{N}\sum_{j=1}^N (\sum_{i=1}^N(w_{ij}\cdot |c_{ij}|) / \|\c_j\|_1),
	%\end{equation}
	where $w_{ij} \in \{0, 1\}$ is the ground truth affinity that takes value $1$ when $\x_i$ and $\x_j$ are from the same subspace and $0$ otherwise.
	SPR takes values in the range of $[0, 1]$, and $\text{SPR} = 1$ if and only if $\C$ is subspace-preserving.
	In addition, we also report the clustering accuracy (ACC) of subspace clustering, which is defined as
	$\max \limits_\pi \frac{1}{N}\sum_{j=1}^N 1_{\{\pi(\p_j) = \hat \p_j\}}$,
	%\begin{align}
	%\text{ACC} = \max \limits_\pi \frac{1}{N}\sum_{j=1}^N 1_{\{\pi(\p_j) = \hat \p_j\}},
	%\end{align}
	where $\p, \hat\p \in \{1,\cdots, n\}^N$ are the groundtruth and estimated assignment of the columns in $X$ to the $n$ subspaces, and $\pi$ is the set of all permutations of $n$ groups.
	
	\begin{figure}[t]
		\begin{center}
			\subfigure[\label{fig:spr-asc-synthetic-data}SPR vs. $D$]{\includegraphics[width=0.225\textwidth,trim={0.5cm 1.2cm 1.1cm 0.45cm},clip]{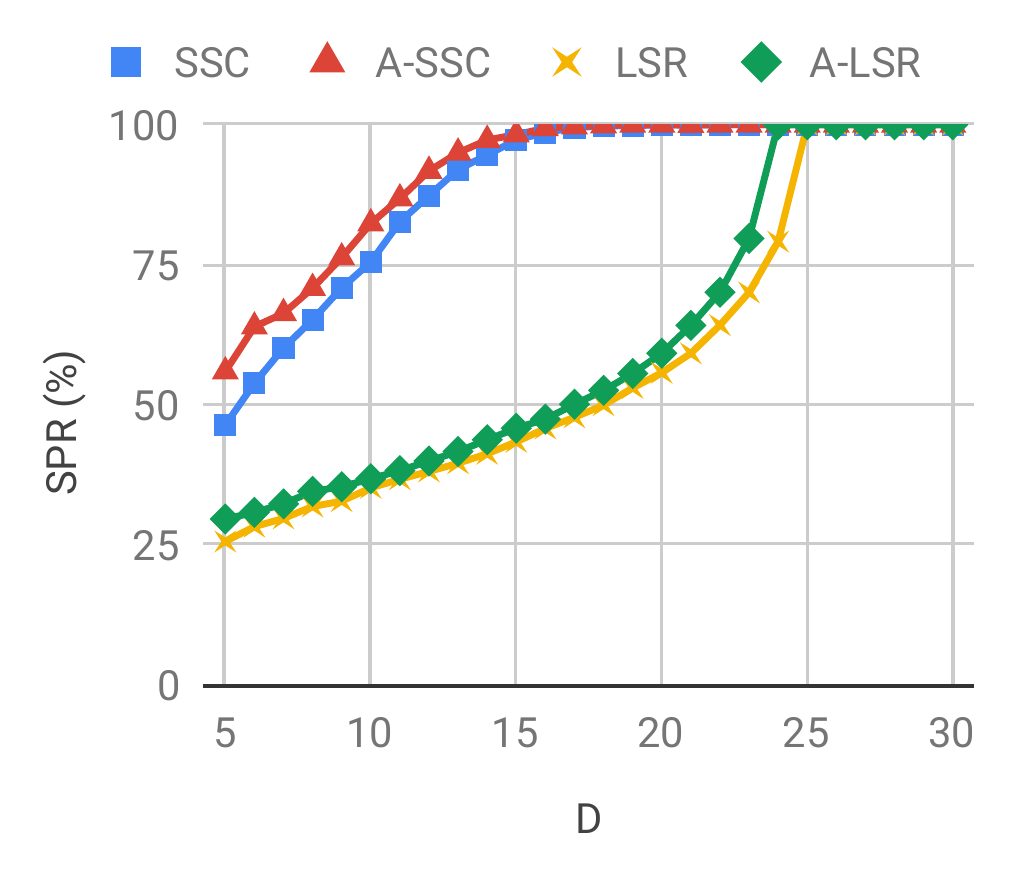}}
			~~
			\subfigure[\label{fig:acc-asc-synthetic-data}ACC vs. $D$]{\includegraphics[width=0.225\textwidth,trim={0.5cm 1.2cm 1.1cm 0.45cm},clip]{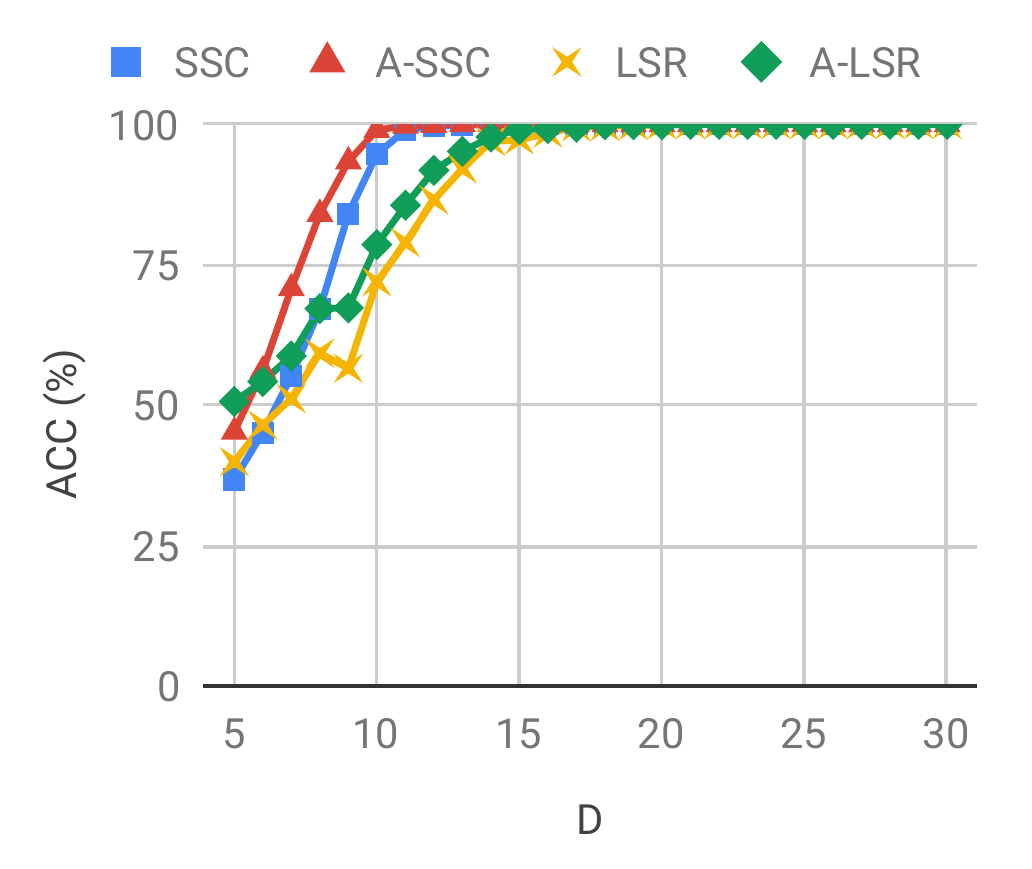}}
			%\centerline{\includegraphics[width=\columnwidth]{}}
			\caption{Performance evaluation on synthetic data. $5$ affine subspaces of dimension $4$ are generated according to the random subspace model, and $20$ points are randomly sampled on each affine subspace. The ambient dimension $D$ is varied on the x-axis. The results are averages over $20$ independent trials.}
			\label{fig:acc-spr-asc-synthetic-data}
		\end{center}
		\vskip -0.2in
	\end{figure}

	The results in our experiments are reported in Figure~\ref{fig:acc-spr-asc-synthetic-data}.
	From Figure~\ref{fig:spr-asc-synthetic-data} we see that A-LSR and LSR produce subspace-preserving solutions when $D$ satisfies the conditions specified in %Corollary
	Theorem~\ref{thm:affine-subspace-clustering-under-random-corollary}\ref{itm:with-affine} and %Corollary
	Theorem~\ref{thm:affine-subspace-clustering-under-random-corollary}\ref{itm:no-affine}, respectively, thus verifying the correctness of these two results.
	Moreover, the solutions are not subspace-preserving as soon as the conditions in %Corollary
	Theorem~\ref{thm:affine-subspace-clustering-under-random-corollary} are violated. That is, A-LSR and LSR do not give subspace-preserving solution whenever $D < \sum_{\ell=1}^{n}d_\ell +n -1 = 24$ and whenever $D < \sum_{\ell=1}^{n}d_\ell +n = 25$, respectively.
	This indicates that these two conditions are tight.
	In addition, the gap between the curves for A-LSR and LSR when $D < 25$, both in terms of SPR (see Figure~\ref{fig:spr-asc-synthetic-data}) and ACC (see Figure~\ref{fig:acc-asc-synthetic-data}), indicates that the affine constraint in A-LSR does play an important role in boosting the performance when the dimension of the ambient space is relatively low.
	
	The range of $D$ for which subspace-preservation is achieved by A-SSC and SSC, on the other hand, extends to significantly smaller values than those that are predicted by %Corollary
	Theorem~\ref{thm:affine-subspace-clustering-under-random-corollary}\ref{itm:with-affine} and %Corollary
	Theorem~\ref{thm:affine-subspace-clustering-under-random-corollary}\ref{itm:no-affine}, respectively, indicating the possibility of deriving tighter bounds for these methods by exploiting special properties of the $\ell_1$ regularizer.\footnote{The work \cite{Li:JSTSP18} presents a theoretical study that is dedicated to A-SSC, but that work only considers the deterministic case and does not provide such a bound under a random subspace model.}
	%Nonetheless, we still observe the pattern consistent with that of LSR and A-LSR that the affine constraint in A-SSC only improve performance in terms of SPR and ACC when the ambient space is low-dimensional.
	Nonetheless, we still observe a pattern that is consistent with that for A-LSR and LSR, namely that the affine constraint in A-SSC improves the performance in terms of SPR and ACC when the ambient space is low-dimensional.

	\subsection{Experiments on real data}
	
	\begin{figure}[t]
		\begin{center}
			\subfigure[\label{fig:real-data-ssc-hopkins155}(A-)SSC on Hopkins 155]{\includegraphics[width=0.225\textwidth,trim={0.2cm 0.0cm 0.2cm 0.3cm},clip]{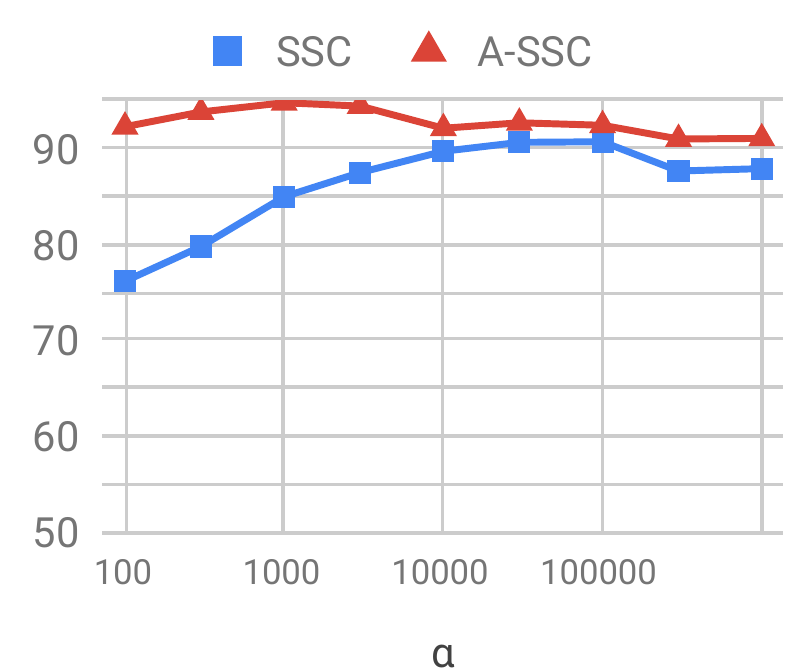}}
			~~
			\subfigure[\label{fig:real-data-lsr-hopkins155}(A-)LSR on Hopkins 155]{\includegraphics[width=0.225\textwidth,trim={0.2cm 0.0cm 0.2cm 0.3cm},clip]{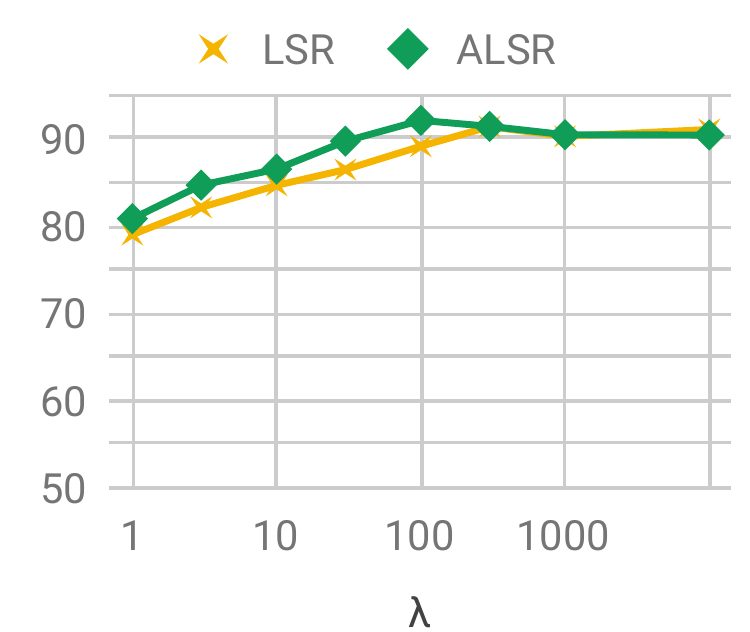}}
			\\
			\subfigure[\label{fig:real-data-ssc-mnist}(A-)SSC on MNIST]{\includegraphics[width=0.225\textwidth,trim={0.2cm 0.0cm 0.2cm 0.3cm},clip]{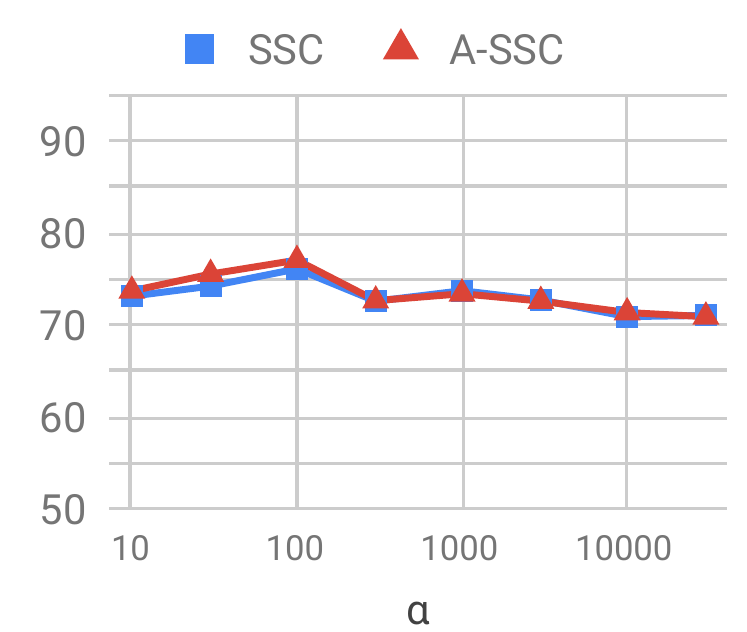}}
			~~
			\subfigure[\label{fig:real-data-lsr-mnist}(A-)LSR on MNIST]{\includegraphics[width=0.225\textwidth,trim={0.2cm 0.0cm 0.2cm 0.3cm},clip]{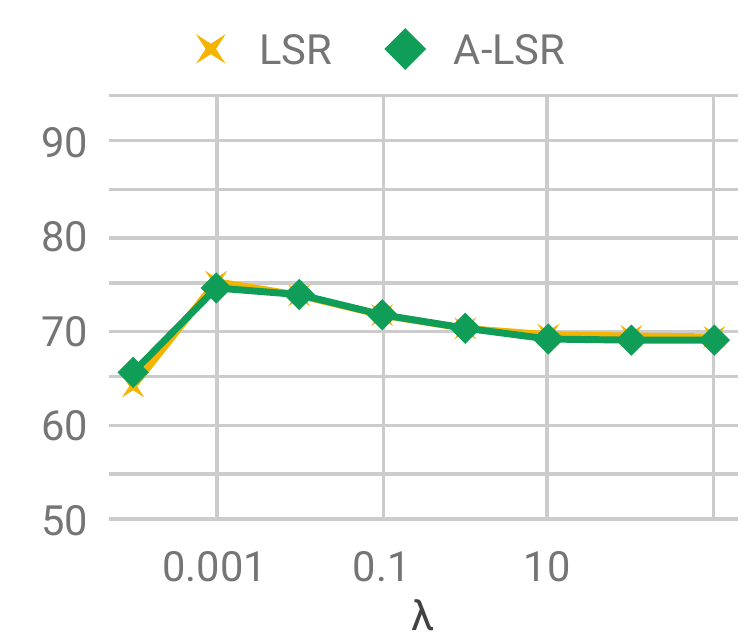}}
			\\
			\subfigure[\label{fig:real-data-ssc-coil}(A-)SSC on Coil-100]{\includegraphics[width=0.225\textwidth,trim={0.2cm 0.0cm 0.2cm 0.3cm},clip]{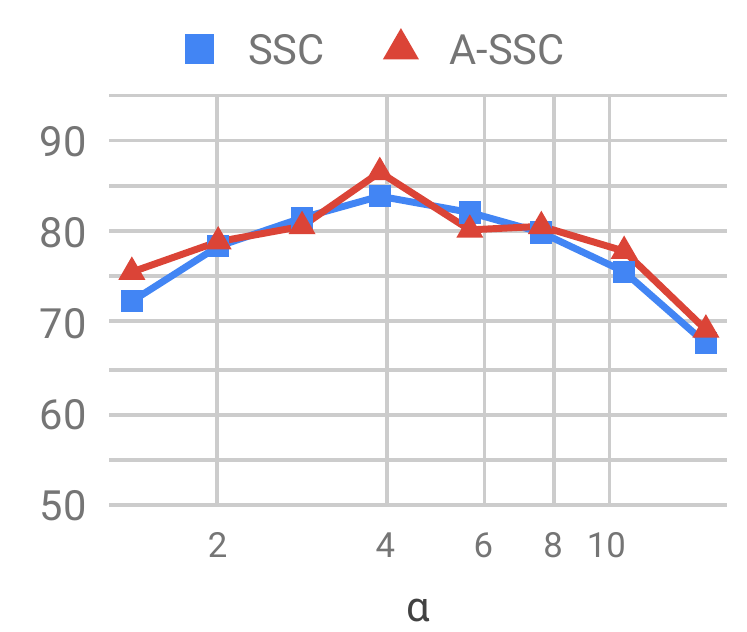}}
			~~
			\subfigure[\label{fig:real-data-lsr-coil}(A-)LSR on Coil-100]{\includegraphics[width=0.23\textwidth,trim={0.2cm 0.0cm 0.2cm 0.3cm},clip]{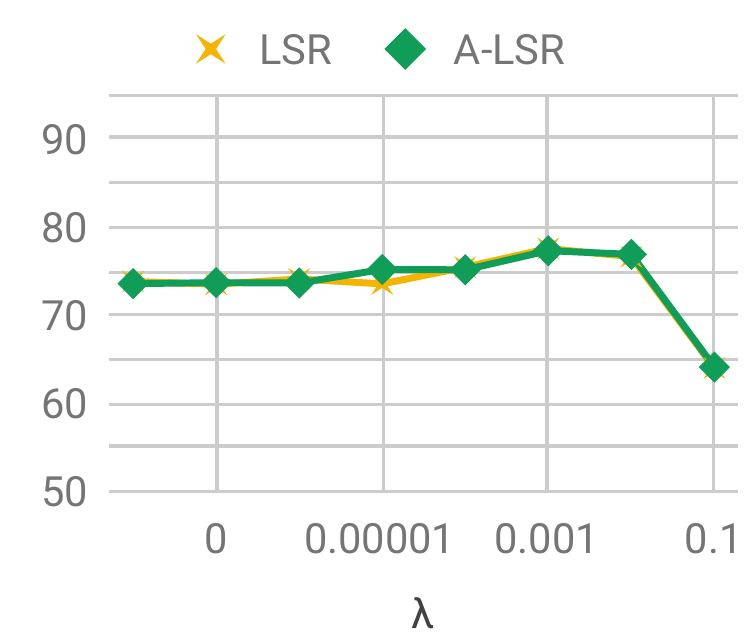}}
			\caption{Performance evaluation on real datasets. We report clustering accuracy (y-axis) versus model parameter value (x-axis). }
			\label{fig:real-data}	\end{center}
		\vskip -0.25in
	\end{figure}
	
	% Is the affine constraint needed for affine subspace clustering on real datasets?
	The literature on subspace clustering usually reports clustering performance of methods with the affine constraint (e.g., \cite{Elhamifar:TPAMI13,Ji:WCACV14}) or without the affine constraint (e.g., \cite{Liu:TPAMI13,Lu:ECCV12,You:CVPR16-EnSC,Lu:ICCV13-TraceLasso,Ji:NIPS17}), thus making it unclear whether the affine constraint is helpful.
	To complement existing studies with the goal of understanding the effect of the affine constraint, we conduct experiments on three commonly used datasets. 
	
	The Hopkins 155 \cite{Tron:CVPR07} is a motion segmentation database that consists of $155$ video sequences with $2$ or $3$ rigid-body motions each.
	%We use all $35$ sequences with $3$ motions and report the average clustering accuracy over the $35$ sequences.
	We report the average clustering accuracy over the $35$ sequences that have $3$ motions.
	The ambient dimension of the data ranges from $30$ to $122$ for different sequences with an average of $57$.
	The MNIST \cite{LeCun:1998} dataset contains $70,000$ images of handwritten digits.
	Each image is of size $32\times 32$.
	Following \cite{You:CVPR16-SSCOMP}, we extract features of dimension $3,\!472$ from each image using the scattering transform network \cite{Bruna:PAMI13} and then project to dimension $500$ via PCA. We randomly choose $1,\!000$ images in each trial to perform clustering and report the average clustering accuracy over $10$ trials.
	The Coil-100 dataset \cite{Nene:1996-coil} contains $7,\!200$ images of $100$ different objects. Each image is of size $128 \times 128$, which is downsampled to size $32 \times 32$ and then concatenated column-wise into a vector of dimension $1,\!024$. We apply mean image subtraction as data preprocessing. We report the average clustering accuracy over $10$ trials where in each trial we pick $10$ classes at random and perform subspace clustering on all images from them.
	%\footnote{Another commonly used database is the Extended Yale B which consists of face images taken under varying illumination conditions.
	%We exclude it from our study as the underlying subspace in that case is not affine.}
	
	The clustering performance of SSC, A-SSC, LSR and A-LSR is reported in Figure~\ref{fig:real-data}.
	We observe from Figure~\ref{fig:real-data-ssc-hopkins155} and \ref{fig:real-data-lsr-hopkins155} that on the Hopkins 155 dataset, A-SSC and A-LSR consistently improve over SSC and LSR, respectively, over a wide range of parameters, indicating the effectiveness of the affine constraint on this dataset.
	This may be explained by the low-dimensionality of the ambient space in which different subspaces (determined by the motions) are not sufficiently separated.
	On the other hand, on both the MNIST and Coil-100 datasets the difference in clustering accuracy between SSC and A-SSC as well as between LSR and A-LSR is very small.
	Specifically, in Figure~\ref{fig:real-data-ssc-mnist}, \ref{fig:real-data-lsr-mnist} and \ref{fig:real-data-lsr-coil} the two curves corresponding to methods with and without the affine constraint are almost overlapping, and in Figure~\ref{fig:real-data-ssc-coil} there is no consistent pattern of any method being better than the other one.
	% This result justifies our claim earlier that the affine constraint may not be needed for high-dimensional data.
	This result confirms our earlier theoretical justification that the affine constraint may not be needed for data with a low intrinsic dimension relative to the dimension of the ambient space.
	
	\begin{figure}[t]
		\begin{center}
			\subfigure{\includegraphics[width=0.225\textwidth,trim={0.2cm 0.00cm 0.2cm 0.35cm},clip]{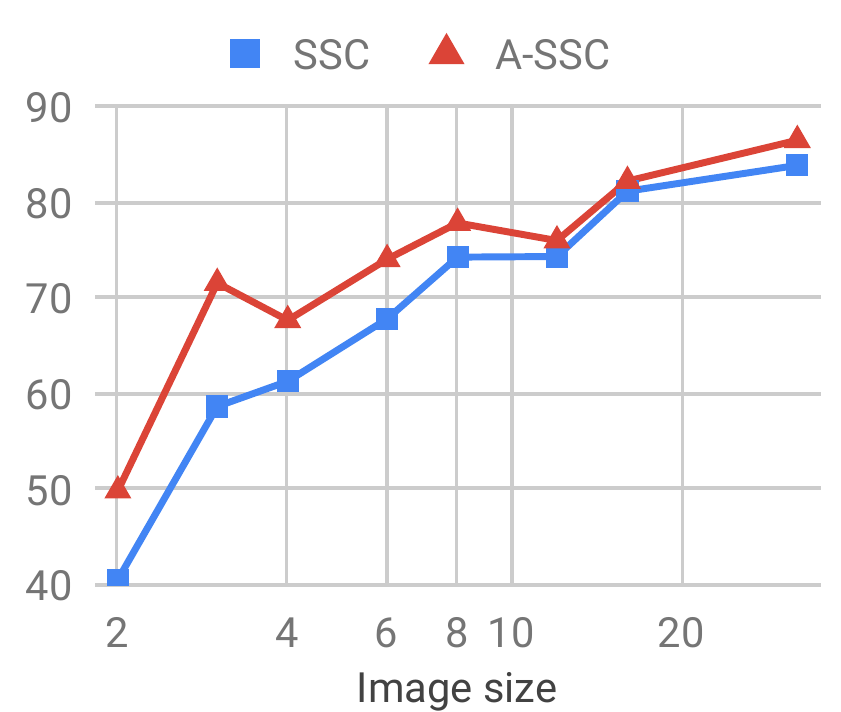}}
			~~
			\subfigure{\includegraphics[width=0.225\textwidth,trim={0.2cm 0.00cm 0.2cm 0.35cm},clip]{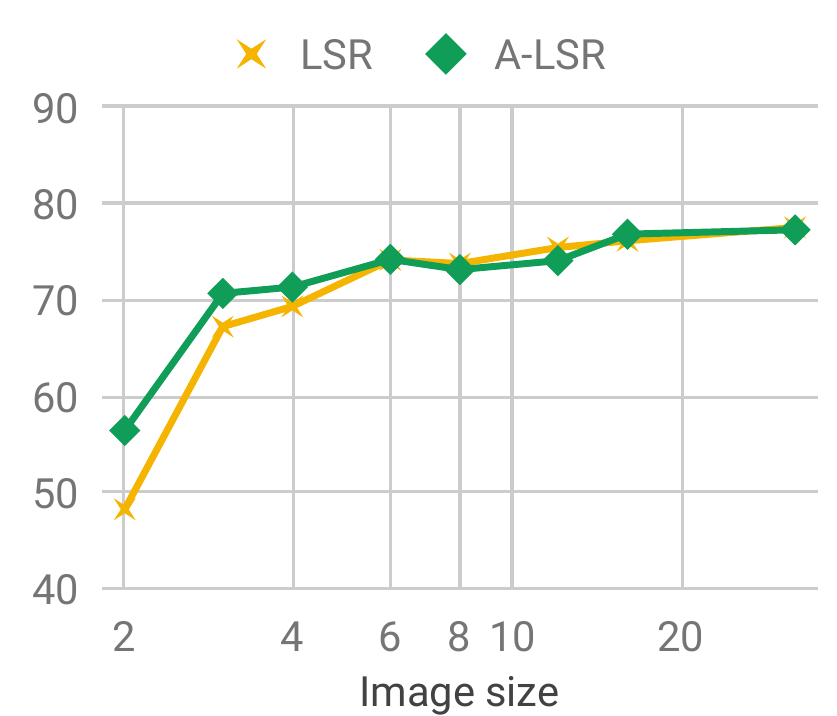}}
			%\centerline{\includegraphics[width=\columnwidth]{}}
			\caption{Performance on Coil-100 with various ambient dimension. Images in Coil-100 are downsampled to size $p$-by-$p$, with $p$ varied on the $x$-axis, and  clustering accuracy shown on the $y$-axis.}
			\label{fig:acc-resolution-real-data}
		\end{center}
		\vskip -0.2in
	\end{figure}
	
	To further evaluate the effect of the ambient dimension, we perform subspace clustering on images from Coil-100 that are downsampled from $32\times32$ to $p \times p$ for $p\in \{2, 3, 4, 6, 8, 12, 16\}$.
	This simulates the effect of varying the ambient dimension caused by varying the image resolution.
	The clustering performance is reported in Figure~\ref{fig:acc-resolution-real-data}.
	The model parameters $\alpha$ (for (A-)SSC) and $\lambda$ (for (A-)LSR) are set to $3.9$ and $0.001$, respectively.
	We can see that the affine constraint in both the cases for SSC and LSR plays a more important role for clustering images with lower resolutions, which are of lower ambient dimension.

	\section{Conclusion}
	\label{sec:conclusion}
	
	%
	%We studied the problem of affine subspace clustering with a focus on understanding the role of the affine constraint in self-expression based subspace clustering methods.
	%Based on our geometric conditions derived in Section~\ref{sec:geometric_results}, we show that the affine constraint may not affect the clustering performance when the ambient dimension is high enough relative to subspace dimensions.
	%This theoretical finding is confirmed by our experiments on synthetic as well as $3$ real datasets that are commonly used in the subspace clustering literature.
	
	We have studied the problem of affine subspace clustering with a focus on understanding the role of the affine constraint in self-expression based subspace clustering methods.
	Based on the geometric conditions derived in Section~\ref{sec:geometric_results},
	%we show that the affine constraint may not affect the clustering performance when the ambient dimension is high enough relative to subspace dimensions.
	we have shown that the affine constraint may have a negligible effect in improving clustering performance when the ambient dimension is large enough relative to the sum of subspace dimensions and the number of subspaces. This theoretical finding was confirmed by our experiments on synthetic data as well as three real datasets commonly used in the subspace clustering literature. Our discovery provides important guidance for practitioners when picking the best  model for their specific subspace clustering tasks.
	
	%% 1.
	%We developed novel geometric conditions which guarantee that the solutions to \eqref{eq:self-representation} and \eqref{eq:affine-self-representation} are subspace-preserving in deterministic setting and random setting, respectively.
	%%
	%% 2.
	%Moreover, we established the equivalence between the notion of \emph{affinely independent} affine subspaces and the independence of the homogenous embedding of the linear subspaces. %Furthermore, we extended the novel geometric conditions which guarantee that the solutions to \eqref{eq:self-representation} and \eqref{eq:affine-self-representation} are subspace-preserving to random setting.
	%%
	%% 3.
	%In addition, we validated our theoretical findings with experiments on synthetic data and real world benchmark data and demonstrated the promise of properly dealing with affine subspace clustering. % theoretically and experimentally.
	
	%\section*{Acknowledgments}
	% Acknowledgements should only appear in the accepted version.
	
	\myparagraph{Acknowledgments.}
	This work has been supported by the National Science Foundation under grants 1618637 and 1704458, and by the Northrop Grumman Corporation's Research in Applications for Learning Machines (REALM) Program. C.-G. Li has been supported by the National Natural Science Foundation of China under grant 61876022 and the Open Project Fund from the Key Laboratory of Machine Perception (MOE), Peking University.
	
	\numberwithin{equation}{section}
	\numberwithin{figure}{section}
	\numberwithin{table}{section}

    \begin{appendices}
	
	The appendices are organized as follows. 
	%In Section~\ref{sec:linear-algebra}, we review the definition of linearly independent subspaces as well as the basics of linear geometry. 
	In Section~\ref{sec:proof}, we present the proofs to the technical results in the paper. 
	In Section~\ref{sec:comparison}, we compare our theoretical results with existing theoretical results for affine subspace clustering. 
	Section~\ref{sec:A_spanA_hA_spanhA} provides a pictorial illustration of several concepts associated with an affine subspace.

	\section{Proofs}
	\label{sec:proof}
	
	In this section, we present proofs for our main theoretical results in Section~\ref{sec:geometric_results} and Section~\ref{sec:random_results}.
	
	\subsection{Technical lemmas}
	We present three lemmas that are used for proving our main theoretical results.
	These lemmas can be established from the definition of span, affine hull and dimension of an (affine) subspace.
	We omit their proofs.  
	
	\begin{lemma}\label{theorem:span-aff}
		Let $\cH \subseteq \RR^D$ be an arbitrary set. We have $\spann(\cH) = \spann(\aff(\cH))$.
	\end{lemma}
	
	%	Following the definition of affine hull we get $\cH \subseteq \aff(\cH)$.
	%	Then, following the definition of span we get  $\spann(\cH) \subseteq \spann(\aff(\cH))$.
	%	
	%	To prove the other direction, we take an arbitrary $\x \in \spann(\aff(\cH))$ and show that $\x \in \spann(\cH)$.	
	%	Let $\{\x_j\}_{j=1}^k$ be an affine basis of $\aff(\cH)$ with the property that $\{\x_j\}_{j=1}^k \subseteq \cH$.
	%	Since $\x \in \spann(\aff(\cH))$, we can express $\x$ as a linear combination of a set of points from $\aff(\cH)$. In addition, each point in this set in turn can be expressed as an affine combination of the points in $\{\x_j\}_{j=1}^k$. Therefore, $\x$ can be expressed as a linear combination of points in $\{\x_j\}_{j=1}^k$.
	%	This shows that $\x \in \spann(\cH)$, i.e., $\spann(\aff(\cH)) \subseteq \spann(\cH)$.

	\begin{lemma}\label{theorem:span-sum}
		Let $\{\cH_\ell \subseteq \RR^D\}_{\ell=1}^m$ be an arbitrary collection of sets. We have 
		\begin{equation}
		\begin{split}
			\spann(\cup_{\ell=1}^m \cH_\ell) &= \spann(\cup_{\ell=1}^m \spann(\cH_\ell)), ~\text{and}~\\
			\aff(\cup_{\ell=1}^m \cH_\ell) &= \aff(\cup_{\ell=1}^m \aff(\cH_\ell)).
		\end{split}
		\end{equation}
	\end{lemma}

	\begin{lemma}\label{thm:dim-affine-vs-span}
		Let $\cA$ be an arbitrary affine subspace. If $\0 \in \cA$ (i.e., if $\cA$ is a linear subspace), then $\dim(\spann(\cA)) = \dim(\cA)$; otherwise, $\dim(\spann(\cA)) = \dim(\cA) + 1$.
	\end{lemma}

	\subsection{Proof of Lemma~\ref{thm:dim-affine-vs-embed-span}}
	
	\begin{proof}
		Let $\cA$ be an affine subspace and $\{\x_j\}_{j=1}^m$ be an affine basis of $\cA$.
		It can be verified from definition that $\hbar(\cA)$ is an affine subspace and that $\{\hbar(\x_j)\}_{j=1}^m$ is an affine basis of $\hbar(\cA)$.
		It follows that $\dim(\hbar(\cA)) = \dim(\cA)=m-1$.
		
		From Lemma~\ref{thm:dim-affine-vs-span} and the fact that $\0 \notin \hbar(\cA)$, we have $\dim(\spann(\hbar(\cA)) = \dim(\hbar(\cA)) + 1 = \dim(\cA)+1$. 	
	\end{proof}
	
	\subsection{Proof of Lemma~\ref{theorem:independence-equivalency}}
	
	\begin{proof}
		From Definition~\ref{def:independent-subspaces}, the collection of subspaces $\{\spann(\hbar(\cA_{\ell}))\}_{\ell=1}^n$ is linearly independent if and only if
		\begin{equation}\label{eq:prf-independent-embedded-subspace}
		\dim(\spann(\cup_{\ell = 1}^n \spann(\hbar(\cA_{\ell})))) = \sum_{\ell=1}^n \dim(\spann(\hbar(\cA_{\ell}))).
		\end{equation}
		From Definition~\ref{def:affinely-independent-affine-subspaces}, 
		the collection of subspaces $\{\cA_{\ell}\}_{\ell=1}^n$ is affinely independent if and only if
		\begin{equation}\label{eq:prf-affinely-independent-affine-subspaces}
		\dim(\aff(\cup_{\ell=1}^{n}\cA_{\ell}))+1 = \sum_{\ell=1}^n \dim(\cA_{\ell})+ n.
		\end{equation}
		To prove the lemma, we only need to show that \eqref{eq:prf-independent-embedded-subspace} and \eqref{eq:prf-affinely-independent-affine-subspaces} are equivalent.
		Note that
		\begin{equation}\label{eq:prf-independence-lhs}
		\begin{split}
		~&\dim(\spann(\cup_{\ell = 1}^n \spann(\hbar(\cA_{\ell}))))\\
		=~&\dim(\spann(\cup_{\ell=1}^{n}\hbar(\cA_{\ell}))) ~~~~~~~~~\text{(by Lemma~\ref{theorem:span-sum})}\\
		=~&\dim(\spann(\hbar(\cup_{\ell=1}^{n}\cA_{\ell})))\\
		=~&\dim(\spann(\aff(\hbar(\cup_{\ell=1}^{n}\cA_{\ell}))))~~\text{(by Lemma~\ref{theorem:span-aff})}\\
		=~&\dim(\spann(\hbar(\aff(\cup_{\ell=1}^{n}\cA_{\ell}))))\\
		=~&\dim(\aff(\cup_{\ell=1}^{n}\cA_{\ell}))+1 ~~~~~~~~~~\text{(by Lemma~\ref{thm:dim-affine-vs-embed-span})}.
		\end{split}
		\end{equation}
		%$\hbar(\aff(\cH)) = \aff(\hbar(\cH))$
		In addition, from Lemma~\ref{thm:dim-affine-vs-embed-span} we have
		\begin{equation}\label{eq:prf-independence-rhs}
		\sum_{\ell=1}^n \dim(\spann(\hbar(\cA_{\ell})))
		= \sum_{\ell=1}^n \dim(\cA_{\ell})+ n.
		\end{equation}
		It follows from \eqref{eq:prf-independence-lhs} and \eqref{eq:prf-independence-rhs} that \eqref{eq:prf-independent-embedded-subspace} and \eqref{eq:prf-affinely-independent-affine-subspaces} are equivalent.
	\end{proof}
	
	\subsection{Proof of Proposition~\ref{thm:condition-2-dimension-bound}}
	
	\begin{proof}
		From Definition~\ref{def:affinely-independent-affine-subspaces}, Lemma~\ref{thm:dim-affine-vs-span} and Lemma~\ref{theorem:span-aff} we have
		\begin{multline*}
		\sum_{\ell=1}^n\dim(\cA_{\ell}) + n - 1 = \dim(\aff(\cup_{\ell=1}^{n}\cA_{\ell})) \\
		\le \dim(\spann(\aff(\cup_{\ell = 1}^n \cA_{\ell}))) = \dim(\spann(\cup_{\ell=1}^{n}\cA_{\ell})),
		\end{multline*}
		which finishes the proof. 
	\end{proof}

	\subsection{Proof of Lemma~\ref{thm:relation-span-independence-affine-independence}}
	\begin{proof}
		%We first prove the ``only if'' direction.
		From Definition~\ref{def:independent-subspaces}, $\{\spann(\cA_{\ell})\}_{\ell=1}^n$ is linearly independent if and only if
		\begin{equation}\label{eq:prf-span-independent}
		\sum_{\ell=1}^n \dim(\spann(\cA_{\ell})) = \dim(\spann(\cup_{\ell=1}^n \spann(\cA_{\ell}))).
		\end{equation}
		Furthermore, from Lemma~\ref{thm:dim-affine-vs-span}, Lemma~\ref{theorem:span-sum} and the fact that $\0 \notin \cup_{\ell=1}^n \cA_{\ell}$,  \eqref{eq:prf-span-independent} holds if and only if
		\begin{equation}\label{eq:prf-span-independent-equivalence}
		\sum_{\ell=1}^n \dim(\cA_{\ell}) + n = \dim(\spann(\cup_{\ell=1}^n \cA_{\ell})).
		\end{equation}
		Therefore, we only need to show that \eqref{eq:prf-span-independent-equivalence} holds if and only if both 1) $\{\cA_{\ell}\}_{\ell=1}^n$ is affinely independent and 2) $\0 \notin \aff(\cup_{\ell=1}^n \cA_{\ell})$.
		
		To prove the ``only if'' direction, suppose that \eqref{eq:prf-span-independent-equivalence} holds.
		From \eqref{eq:affine-dim-bound}, \eqref{eq:prf-span-independent-equivalence}, Lemma~\ref{theorem:span-aff} and Lemma~\ref{thm:dim-affine-vs-span}, we have
		\begin{equation}\label{eq:prf-dim-inequalities}
		\begin{split}
		\dim(\aff(\cup_{\ell=1}^{n}\cA_{\ell})) + 1
		&\le \sum_{\ell=1}^n\dim(\cA_{\ell}) + n\\
		&= \dim(\spann(\cup_{\ell=1}^n \cA_{\ell}))\\
		&= \dim(\spann(\aff(\cup_{\ell=1}^n \cA_{\ell})))\\
		&\le \dim(\aff(\cup_{\ell=1}^n \cA_{\ell})) + 1.
		\end{split}
		\end{equation}
		It follows that equality is achieved for all inequalities in \eqref{eq:prf-dim-inequalities}.
		In particular, equality in the first line of \eqref{eq:prf-dim-inequalities} implies that $\{\cA_{\ell}\}_{\ell=1}^n$ is affinely independent, and equality in the last line of \eqref{eq:prf-dim-inequalities} implies that $\0 \notin \aff(\cup_{\ell=1}^n \cA_{\ell})$.
		
		To prove the ``if'' direction, we assume that $\{\cA_{\ell}\}_{\ell=1}^n$ is affinely independent and that $\0 \notin \aff(\cup_{\ell=1}^n \cA_{\ell})$.
		By Definition~\ref{def:independent-subspaces}, Lemma~\ref{thm:dim-affine-vs-span} and Lemma~\ref{theorem:span-aff}, we have
		\begin{equation}
		\begin{split}
		\sum_{\ell=1}^n\dim(\cA_{\ell}) + n &= \dim(\aff(\cup_{\ell=1}^{n}\cA_{\ell})) + 1\\
		&=\dim(\spann(\aff(\cup_{\ell=1}^n \cA_{\ell})))\\
		&=\dim(\spann(\cup_{\ell=1}^n \cA_{\ell})),
		\end{split}	
		\end{equation}
		therefore \eqref{eq:prf-span-independent-equivalence} holds. This completes the proof.
	\end{proof}

	\subsection{Proof of Proposition~\ref{thm:condition-1-dimension-bound}}
	
	\begin{proof}
		From the proof for the ``if'' direction of Lemma~\ref{thm:relation-span-independence-affine-independence}, if $\{\cA_{\ell}\}_{\ell=1}^n$ is affinely independent and $\0 \notin \aff(\cup_{\ell=1}^n \cA_{\ell})$ then \eqref{eq:prf-span-independent-equivalence} holds, which is the desired result. 
	\end{proof}

	\subsection{Proof of Lemma~\ref{thm:affine-subspace-clustering-under-random}}
	
	\begin{proof}
		%Let $\{\w_{0,\ell}, \w_{1,\ell}, \cdots, \w_{d_\ell,\ell}\}_{\ell=1}^n$ be a set of vectors drawn independently and uniformly at random from the unit sphere of $\RR^D$, and 
		%let $\cA_{\ell} = \w_{0,\ell}+\spann\{\w_{1,\ell}, \cdots, \w_{d_\ell,\ell}\}$ for each $\ell=1, \cdots, n$. 
		%We need to show that the collection of affine subspaces $\{\cA_{\ell}\}_{\ell=1}^n$ is affinely independent with probability $1$.
		
		For each $\ell \in \{1, \cdots, n\}$, let $\tilde{\w}_{i, \ell} := \w_{i,\ell}$ for $i=0$ and $\tilde{\w}_{i, \ell} := \w_{0,\ell} + \w_{i,\ell}$ for $i\in \{1, \cdots, d_\ell\}$. 
		In addition, let $\widetilde{\cW} = \cup_{\ell=1}^n \widetilde{\cW}_\ell$, where
		\begin{equation}\label{eq:prf-def-W}
			\widetilde{\cW}_\ell = \{\tilde{\w}_{0,\ell}, \tilde{\w}_{1,\ell}, \cdots, \tilde{\w}_{d_\ell,\ell}\}.
		\end{equation}
		Note that $\aff(\widetilde{\cW}_\ell) = \cA_{\ell}$. By using Lemma~\ref{theorem:span-sum}, we have
		% It is easy to verify that the set $\tilde{\cW}_\ell := \{\tilde{\w}_{0,\ell}, \tilde{\w}_{1,\ell}, \cdots, \tilde{\w}_{d_\ell,\ell}\}$ is an affine basis of $\cA_{\ell}$ with probability $1$ for each $\ell$.
		\begin{multline}\label{eq:prf-W-inclusion}
		\aff(\widetilde{\cW}) = \aff(\cup_{\ell = 1}^n \widetilde{\cW}_\ell)\\
		= \aff(\cup_{\ell = 1}^n \aff(\widetilde{\cW}_\ell)) = \aff(\cup_{\ell = 1}^n\cA_\ell).
		\end{multline}	
		Meanwhile, note that
		\begin{equation}\label{eq:prf-W-cardinality}
		\card(\widetilde{\cW}) = \sum_{\ell=1}^n d_\ell + n ~~ \text{with probability}~~ 1.
		\end{equation} 
		Combining \eqref{eq:prf-W-cardinality} with $D \ge \sum_{\ell=1}^n d_\ell+n - 1$, we have
		\begin{equation}\label{eq:prf-W-independent}
		\widetilde{\cW} ~\text{is affinely independent with probability}~1.
		\end{equation}
		It follows from \eqref{eq:prf-W-inclusion}, \eqref{eq:prf-W-cardinality} and \eqref{eq:prf-W-independent} that 
		\begin{multline}\label{eq:prf-random}
		\dim(\aff(\cup_{\ell = 1}^n\cA_\ell)) = \dim(\aff(\widetilde{\cW})) = \card(\widetilde{\cW}) - 1 \\
		= \sum_{\ell=1}^n d_\ell + n - 1 = \sum_{\ell=1}^n \dim(\cA_{\ell}) + n - 1,
		\end{multline}
		with probability $1$. 
		Combining this with Definition~\ref{def:affinely-independent-affine-subspaces} completes the proof. 
	\end{proof}
	
	\subsection{Proof of Lemma~\ref{thm:subspace-clustering-under-random}}

	\begin{proof}
		%Let $\{\w_{0,\ell}, \w_{1,\ell}, \cdots, \w_{d_\ell,\ell}\}_{\ell=1}^n$ be a set of vectors drawn independently and uniformly at random from the unit sphere of $\RR^D$, and 
		%let $\cA_{\ell} = \w_{0,\ell}+\spann\{\w_{1,\ell}, \cdots, \w_{d_\ell,\ell}\}$ for each $\ell=1, \cdots, n$. 
		From Lemma~\ref{thm:affine-subspace-clustering-under-random}, we know that $\{\cA_{\ell}\}_{\ell=1}^n$ is affinely independent with probability $1$.
		In the following, we show that $\0 \notin \aff(\cup_{\ell=1}^n \cA_{\ell})$ with probability $1$.
		
		Let $\widetilde{\cW}, \widetilde{\cW}_\ell$ be defined as in the proof to Lemma~\ref{thm:affine-subspace-clustering-under-random}. 
		Following the same argument as before we know that \eqref{eq:prf-W-inclusion}, \eqref{eq:prf-W-cardinality}, \eqref{eq:prf-W-independent} and \eqref{eq:prf-random} hold. 
		In addition, note that
		\begin{multline}\label{eq:prf-W-span}
		\spann(\widetilde{\cW}) = \spann(\cup_{\ell = 1}^n\widetilde{\cW}_\ell) \\
		\subseteq \spann(\cup_{\ell = 1}^n\aff(\widetilde{\cW}_\ell)) = \spann(\cup_{\ell = 1}^n\cA_\ell).
		\end{multline}
		Meanwhile, combining \eqref{eq:prf-W-cardinality} with $D \ge \sum_{\ell=1}^n d_\ell+n$ gives
		\begin{equation}\label{eq:prf-W-linear-independent}
		\widetilde{\cW} ~\text{is linearly independent with probability}~1.
		\end{equation}
		It follows from \eqref{eq:prf-W-span} and \eqref{eq:prf-W-linear-independent} that 
		\begin{equation}\label{eq:prf-random-linear}
		\dim(\spann(\cup_{\ell = 1}^n\cA_\ell)) \ge \dim(\spann(\widetilde{\cW}))= \text{card}(\widetilde{\cW}),
		\end{equation}
		with probability $1$. 
		By using Lemma~\ref{theorem:span-aff}, \eqref{eq:prf-random-linear} and \eqref{eq:prf-random} we get
		\begin{multline}
		\dim(\spann(\aff(\cup_{\ell = 1}^n\cA_\ell))) = \dim(\spann(\cup_{\ell = 1}^n\cA_\ell)) \\
		\ge \text{card}(\widetilde{\cW}) = \dim(\aff(\cup_{\ell = 1}^n\cA_\ell)) + 1,
		\end{multline}
		with probability $1$. 
		Combining this with Lemma~\ref{thm:dim-affine-vs-span} shows that $\0 \notin \aff(\cup_{\ell=1}^n \cA_{\ell})$, as desired.
	\end{proof}
	
	\section{Comparison with Existing Results}
	\label{sec:comparison}
	
	We compare our technical results with those in existing works. 
	To the best of our knowledge, the only papers that have provided theoretical guarantees for the correctness of affine subspace methods are \cite{Elhamifar:CVPR09,Li:JSTSP18} which study the affine sparse subspace clustering (i.e., A-SSC), and \cite{Tsakiris:AffinePAMI17} which study the affine algebraic subspace clustering (i.e., A-ASC). 
	
	\subsection{Comparison with results for A-SSC}
	
	A-SSC refers to the method in \eqref{eq:affine-self-representation} with $f(\cdot)=\|\cdot\|_1$ and $\Omega = \{\C: \diag(\C)=\0\}$. 
	It is originally proposed in \cite{Elhamifar:CVPR09}, where a theoretical analysis of A-SSC is provided that states that A-SSC produces subspace-preserving solutions if the collection of subspaces $\{\spann(\hbar(\cA_{\ell}))\}_{\ell=1}^n$ is linearly independent. 
	This result from \cite{Elhamifar:CVPR09} is a special case of Lemma~\ref{thm:affine-subspace-clustering-on-affine-EBD}. 
	As stated in the discussion following Lemma~\ref{thm:affine-subspace-clustering-on-affine-EBD}, such a result does not provide clear geometric insights as it is characterized by the embedding of the affine subspaces in the homogeneously embedded ambient space. 
	
	In a recent work \cite{Li:JSTSP18} that provides a dedicated analysis of A-SSC, it is shown that A-SSC provides subspace-preserving solutions if the collection of affine subspaces $\{\cA_{\ell}\}_{\ell=1}^n$ is affinely independent. 
	This result from \cite{Li:JSTSP18} is a special case of Corollary~\ref{thm:affine-subspace-clustering-on-affine-affinely-independent}. 
	In addition, \cite{Li:JSTSP18} also presents conditions that guarantee subspace-preserving property of A-SSC beyond the affinely independent assumption. 
	However, the derivation of these conditions rely heavily on the property of $f(\cdot)=\|\cdot\|_1$ in A-SSC and do not apply to the general formulation in \eqref{eq:affine-self-representation} for a broad range of $f(\cdot)$. 
	
	The relation between the two previous condition in \cite{Elhamifar:CVPR09} and \cite{Li:JSTSP18} that $\{\spann(\hbar(\cA_{\ell}))\}_{\ell=1}^n$ is linearly independent and that $\{\cA_{\ell}\}_{\ell=1}^n$ is affinely independent has been unclear, making it hard to compare these two results. 
	This has now been clarified by our Theorem~\ref{theorem:independence-equivalency}, which shows that these two conditions are equivalent to each other.

	\subsection{Comparison with results for A-ASC}
	
	The algebraic subspace clustering (ASC) method, also known as the generalized principal component analysis \cite{Vidal:PAMI05,Tsakiris:FSASCICCV15,Tsakiris:SIAM17}, is a subspace clustering method based on fitting the data with a system of homogeneous polynomials followed by factorization of the polynomials. 
	To handle data drawn from a union of affine subspaces, the A-ASC first computes the homogeneous embedding of the data, then apply ASC to the embedded data. 
	A recent work \cite{Tsakiris:AffinePAMI17} establishes conditions under which A-ASC is guaranteed to produce correct clustering. 
	In particular, a core condition is that the collection of direction subspaces associated with the affine subspaces is \emph{transversal}, a mild assumption that holds with probability $1$ under the random affine subspace model in Definition~\ref{def:random-affine-subspace-model} as long as $d_\ell < D$ for each $\ell=1, \cdots, n$ (\cite[Proposition 3]{Tsakiris:AffinePAMI17}). 
	A comparison of this result from \cite{Tsakiris:AffinePAMI17} with the results in Theorem~\ref{thm:affine-subspace-clustering-under-random-corollary} suggests that A-ASC may work for a broader range of problems than methods based on \eqref{eq:affine-self-representation} and \eqref{eq:self-representation}. 
	Nonetheless, the computational complexity of A-ASC is exponentially large in the dimension of the ambient space, making it only applicable to low-dimensional data sets.

	\section{Relation Between $\cA$, $\spann(\cA)$, $\hbar (\cA)$, and $\spann(\hbar(\cA))$ via an Example}
	\label{sec:A_spanA_hA_spanhA}
	
	In Figure~\ref{fig:A-spanA-hA-spanhA}, we provide an example that illustrates three concepts associated with an affine subspace $\cA$: the linear subspace $\spann(\cA)$, the homogeneous embedding $\hbar (\cA)$, and the linear subspace $\spann(\hbar(\cA))$. Note that the linear subspace $\spann(\cA)$ is the entire plane of $\RR^2$ and the linear subspace $\spann(\hbar(\cA))$ is a subspace of $\RR^3$.
	
	\begin{figure}[ht]
		%	\vskip 0.2in
		\begin{center}
			\subfigure[\label{fig:affine-subspace-A}]{
				\begin{tikzpicture}
				[scale=1.45,axis/.style={thin, ->, >=stealth'}] %very thick
				% axis
				\draw[axis] (xyz cs:x=-1.1) -- (xyz cs:x=1.1) node[right] {$^{x}$};
				\draw[axis] (xyz cs:y=-0.5) -- (xyz cs:y=1.1) node[right] {$^{y}$};
				%\draw[axis] (xyz cs:z=-1.1) -- (xyz cs:z=1.1);
				%\path[draw=red] (0,0) -- (1,1) -- (2,1) circle (10pt);
				\draw[very thick, blue] (xyz cs:y=0.7, x=-0.2) -- (xyz cs:y=-0.2,x=0.7) node[right] {$^{\cA_1}$}; % 0.5-x
				%\draw[very thick, red] (xyz cs:y=0.25, z=-1) -- (xyz cs:y=0.25,z=1) node[left] {$\cA_2$};
				\end{tikzpicture}
			}
			~
			\subfigure[\label{fig:span-of-affine-subspace-A}]{
				\begin{tikzpicture}[
				scale=1.45,
				axis/.style={thin, ->, >=stealth'}
				]
				\draw[axis] (xyz cs:x=-1.1) -- (xyz cs:x=1.1) node[right] {$^{x}$};
				\draw[axis] (xyz cs:y=-0.5) -- (xyz cs:y=1.1) node[right] {$^{y}$};
				%\draw[axis] (xyz cs:z=-1.1) -- (xyz cs:z=1.1);
				\shadedraw[fill=green!20!white, draw=green!50!black](0.7,-0.2) -- (-0.2, 0.7)-- (0,0);
				\draw[very thick, blue] (xyz cs:y=0.7, x=-0.2) -- (xyz cs:y=-0.2,x=0.7) node[right] {$^{\spann(\cA_1)}$}; % 0.5-x
				%\draw[very thick, red] (xyz cs:y=0.5,z=-1) -- (xyz cs:y=0.5,z=1) node[left] {$\cA_2$};
				\fill[blue] (xyz cs:x=0,y=0) circle (1.5pt) 	node[left] {$^{^{(0,0)}} $};
				%\filldraw[fill=green!10!white, draw=green!10!](0,0) (-1.1,-0.5) rectangle (1.1,1.1);
				%\fill[pink] (1,0) -- (1,1) arc [start angle=90, end angle=180, radius=1] -- (0,0);
				%\fill[pink!] (1,0) -- (1,1) arc [start angle=90, end angle=180, radius=1] -- (0,0);
				\end{tikzpicture}
			}
			\\
			\subfigure[\label{fig:HE-of-affine-subspace-A}]
			{\begin{tikzpicture}[
				scale=1.45,
				axis/.style={thin, ->, >=stealth'}
				]
				% axis
				\draw[axis] (xyz cs:x=-1.1) -- (xyz cs:x=1.1) node[right] {$^{y}$};
				\draw[axis] (xyz cs:y=-0.5) -- (xyz cs:y=1.1) node[right] {$^{z}$};
				\draw[axis] (xyz cs:z=-1.8) -- (xyz cs:z=1.5) node[left] {$^{x}$};
				\draw[very thick, red] (xyz cs:x=0.7, z=-0.2, y=.5) -- (xyz cs:x=-0.2,z=0.7,y=.5) node[left] {$^{\hbar(\cA_1)}$}; % 0.5-x
				\draw[thick, blue!40!] (xyz cs:x=0.7, z=-0.2) -- (xyz cs:x=-0.2,z=0.7) node[left] {$^{\cA_1}$}; % 0.5-x
				\fill[black] (xyz cs:x=0,y=0.5,z=0) circle (1.5pt) node[right] {$^{^{(0,0,1)}}$};
				%\draw[very thick, blue] (xyz cs:y=0.5,x=-1) -- (xyz cs:y=0.5,x=1) node[above] {$\cA_1$};
				%\fill[red] (xyz cs:z=0.5) circle (2pt)	node[left] {$^{\spann(\cA_1)}$};
				\end{tikzpicture}
			}
			\subfigure[\label{fig:span-of-HE-of-affine-subspace-A}]
			{\begin{tikzpicture}[
				scale=1.45,
				axis/.style={thin, ->, >=stealth'}
				]
				% axis
				%\shadedraw [shading=axis] (0.7,-0.2,0.5) rectangle (-0.2,0.7,.5);
				\draw[axis] (xyz cs:x=-1.1) -- (xyz cs:x=1.1) node[right] {$^{y}$};
				\draw[axis] (xyz cs:y=-0.5) -- (xyz cs:y=1.1) node[right] {$^{z}$};;
				\draw[axis] (xyz cs:z=-1.8) -- (xyz cs:z=1.5) node[left] {$^{x}$};;
				\fill[black] (xyz cs:x=0,y=0.5,z=0) circle (1.5pt) node[left] {$^{^{(0,0,1)}}$};
				%\draw[thick, blue!60!] (xyz cs:x=0.7, z=-0.2) -- (xyz cs:x=-0.2,z=0.7) node[left] {$^{\cA_1}$}; % 0.5-x
				%\draw[very thick, blue] (xyz cs:y=0.5,x=-1) -- (xyz cs:y=0.5,x=1) node[above] {$\cA_1$};
				\shadedraw[fill=green!20!white, draw=green!50!black](0.7,0.5,-0.2) -- (-0.2,0.5, 0.7)-- (0,0,0);
				\fill[red] (xyz cs:x=0,y=0,z=0) circle (1.5pt) node[right] {$^{^{(0,0,0)}} $};
				\draw[very thick, red] (xyz cs:x=0.7, z=-0.2, y=.5) -- (xyz cs:x=-0.2,z=0.7,y=.5) node[left] {$^{\spann (\hbar(\cA_1))}$}; % 0.5-x
				\draw[thick, blue!40!] (xyz cs:x=0.7, z=-0.2) -- (xyz cs:x=-0.2,z=0.7) node[left] {$^{\cA_1}$}; % 0.5-x
				
				\end{tikzpicture}
			}
			\caption{Illustration of the four concepts: affine subspace $\cA$, $\spann(\cA)$, homogenous embedding $\hbar (\cA)$, and $\spann(\hbar (\cA))$.}
			\label{fig:A-spanA-hA-spanhA}
		\end{center}
		\vskip -0.2in
	\end{figure}
	\end{appendices}
	{
		\small
		\bibliographystyle{ieee_fullname}
		\bibliography{biblio/vidal,biblio/vision,biblio/math,biblio/learning,biblio/sparse,biblio/geometry,biblio/dti,biblio/recognition,biblio/surgery,biblio/coding,biblio/segmentation,biblio/dataset}

\begin{thebibliography}{10}\itemsep=-1pt

\bibitem{Bako-Vidal:HSCC08}
Laurent Bako and Ren{\'{e}} Vidal.
\newblock Algebraic identification of {MIMO} {SARX} models.
\newblock In {\em Hybrid Systems: Computation and Control}, pages 43--57.
  Spinger-Verlag, 2008.

\bibitem{Bruna:PAMI13}
Joan Bruna and Stephane Mallat.
\newblock Invariant scattering convolution networks.
\newblock {\em IEEE Trans. Pattern Anal. Mach. Intell.}, 35(8):1872--1886, Aug.
  2013.

\bibitem{Costeira:IJCV98}
Joao~Paulo Costeira and Takeo Kanade.
\newblock A multibody factorization method for independently moving objects.
\newblock {\em International Journal of Computer Vision}, 29(3):159--179, 1998.

\bibitem{Elhamifar:CVPR09}
Ehsan Elhamifar and Ren\'{e} Vidal.
\newblock Sparse subspace clustering.
\newblock In {\em {IEEE} Conference on Computer Vision and Pattern
  Recognition}, pages 2790--2797, 2009.

\bibitem{Elhamifar:TPAMI13}
Ehsan Elhamifar and Ren\'{e} Vidal.
\newblock Sparse subspace clustering: Algorithm, theory, and applications.
\newblock {\em {IEEE} Transactions on Pattern Analysis and Machine
  Intelligence}, 35(11):2765--2781, 2013.

\bibitem{Favaro:CVPR11}
Paolo Favaro, Ren\'{e} Vidal, and Avinash Ravichandran.
\newblock A closed form solution to robust subspace estimation and clustering.
\newblock In {\em IEEE Conference on Computer Vision and Pattern Recognition},
  pages 1801 --1807, 2011.

\bibitem{cvx}
Michael Grant and Stephen Boyd.
\newblock {CVX}: Matlab software for disciplined convex programming, version
  1.21.
\newblock \url{http://cvxr.com/cvx/}, 2011.

\bibitem{Ji:WCACV14}
Pan Ji, Mathieu Salzmann, and Hongdong Li.
\newblock Efficient dense subspace clustering.
\newblock In {\em IEEE Winter Conference on Applications of Computer Vision},
  pages 461--468. IEEE, 2014.

\bibitem{Ji:NIPS17}
Pan Ji, Tong Zhang, Hongdong Li, Mathieu Salzmann, and Ian Reid.
\newblock Deep subspace clustering networks.
\newblock In {\em Advances in Neural Information Processing Systems}, pages
  24--33, 2017.

\bibitem{Jiang:COA18}
Hao Jiang, Daniel~P Robinson, Ren{\'e} Vidal, and Chong You.
\newblock A nonconvex formulation for low rank subspace clustering: algorithms
  and convergence analysis.
\newblock {\em Computational Optimization and Applications}, 70(2):395--418,
  2018.

\bibitem{LeCun:1998}
Yann LeCun, L\'{e}on Bottou, Yoshua Bengio, and Patrick Haffner.
\newblock Gradient-based learning applied to document recognition.
\newblock {\em Proceedings of the IEEE}, 86(11):2278 -- 2324, 1998.

\bibitem{Li:TIP17}
Chun-Guang Li, Chong You, and Ren\'{e} Vidal.
\newblock Structured sparse subspace clustering: A joint affinity learning and
  subspace clustering framework.
\newblock {\em {IEEE} Transactions on Image Processing}, 26(6):2988--3001,
  2017.

\bibitem{Li:JSTSP18}
Chun-Guang Li, Chong You, and Ren\'{e} Vidal.
\newblock On geometric analysis of affine sparse subspace clustering.
\newblock {\em {IEEE} Journal on Selected Topics in Signal Processing},
  12(6):1520--1533, 2018.

\bibitem{Li:CVPR19-subspace}
Yuanman Li, Jiantao Zhou, Xianwei Zheng, Jinyu Tian, and Yuan~Yan Tang.
\newblock Robust subspace clustering with independent and piecewise identically
  distributed noise modeling.
\newblock In {\em Proceedings of the IEEE Conference on Computer Vision and
  Pattern Recognition}, pages 8720--8729, 2019.

\bibitem{Liu:TPAMI13}
Guangcan Liu, Zhouchen Lin, Shuicheng Yan, Ju Sun, and Yi Ma.
\newblock Robust recovery of subspace structures by low-rank representation.
\newblock {\em IEEE Transactions on Pattern Analysis and Machine Intelligence},
  35(1):171--184, 2013.

\bibitem{Liu:ICML10}
Guangcan Liu, Zhouchen Lin, and Yingrui Yu.
\newblock Robust subspace segmentation by low-rank representation.
\newblock In {\em International Conference on Machine Learning}, pages
  663--670, 2010.

\bibitem{Lu:TPAMI18}
Canyi Lu, Jiashi Feng, Zhouchen Lin, Tao Mei, and Shuicheng Yan.
\newblock Subspace clustering by block diagonal representation.
\newblock {\em IEEE Transactions on Pattern Analysis and Machine Intelligence},
  2018.

\bibitem{Lu:ICCV13-TraceLasso}
Can-Yi Lu, Zhouchen Lin, and Shuicheng Yan.
\newblock Correlation adaptive subspace segmentation by trace lasso.
\newblock In {\em {IEEE} International Conference on Computer Vision}, pages
  1345--1352, 2013.

\bibitem{Lu:ECCV12}
Can-Yi Lu, Hai Min, Zhong-Qiu Zhao, Lin Zhu, De-Shuang Huang, and Shuicheng
  Yan.
\newblock Robust and efficient subspace segmentation via least squares
  regression.
\newblock In {\em European Conference on Computer Vision}, pages 347--360,
  2012.

\bibitem{McWilliams:DMKD14}
Brian McWilliams and Giovanni Montana.
\newblock Subspace clustering of high dimensional data: a predictive approach.
\newblock {\em Data Mining and Knowledge Discovery}, 28(3):736--772, 2014.

\bibitem{Nene:1996-coil}
Sameer Nene, Shree~K. Nayar, and Hiroshi Murase.
\newblock Columbia object image library ({COIL}-100).
\newblock {\em Technical Report CUCS-006-96}, 1996.

\bibitem{Pourkamali:arXiv18}
Farhad Pourkamali-Anaraki and Stephen Becker.
\newblock Efficient solvers for sparse subspace clustering.
\newblock {\em arXiv preprint arXiv:1804.06291}, 2018.

\bibitem{Rao:PAMI10}
Shankar Rao, Roberto Tron, Ren\'{e} Vidal, and Yi Ma.
\newblock Motion segmentation in the presence of outlying, incomplete, or
  corrupted trajectories.
\newblock {\em {IEEE} Transactions on Pattern Analysis and Machine
  Intelligence}, 32(10):1832--1845, 2010.

\bibitem{Robinson:arxiv19}
Daniel~P Robinson, Rene Vidal, and Chong You.
\newblock Basis pursuit and orthogonal matching pursuit for subspace-preserving
  recovery: Theoretical analysis.
\newblock {\em arXiv preprint arXiv:1912.13091}, 2019.

\bibitem{Soltanolkotabi:AS12}
Mahdi Soltanolkotabi and Emmanuel~J. Cand\`es.
\newblock A geometric analysis of subspace clustering with outliers.
\newblock {\em Annals of Statistics}, 40(4):2195--2238, 2012.

\bibitem{Soltanolkotabi:AS14}
Mahdi Soltanolkotabi, Ehsan Elhamifar, and Emmanuel~J. Cand\`es.
\newblock Robust subspace clustering.
\newblock {\em Annals of Statistics}, 42(2):669--699, 2014.

\bibitem{Tomasi:IJCV92}
Carlo Tomasi and Takeo Kanade.
\newblock Shape and motion from image streams under orthography.
\newblock {\em International Journal of Computer Vision}, 9(2):137--154, 1992.

\bibitem{Tron:CVPR07}
Roberto Tron and Ren\'{e} Vidal.
\newblock A benchmark for the comparison of 3-{D} motion segmentation
  algorithms.
\newblock In {\em {IEEE} Conference on Computer Vision and Pattern
  Recognition}, pages 1--8, 2007.

\bibitem{Tsakiris:FSASCICCV15}
M.C. Tsakiris and R. Vidal.
\newblock Filtrated spectral algebraic subspace clustering.
\newblock In {\em ICCV Workshop on Robust Subspace Learning and Computer
  Vision}, pages 28--36, 2015.

\bibitem{Tsakiris:AffinePAMI17}
Manolis~C. Tsakiris and Ren\'{e} Vidal.
\newblock Algebraic clustering of affine subspaces.
\newblock {\em {IEEE} Transactions on Pattern Analysis and Machine
  Intelligence}, 40(2):482--489, 2017.

\bibitem{Tsakiris:SIAM17}
M.~C. Tsakiris and R. Vidal.
\newblock Filtrated algebraic subspace clustering.
\newblock {\em {SIAM} Journal on Imaging Sciences}, 10(1):372--415, 2017.

\bibitem{Tsakiris:ICML18}
Manolis~C. Tsakiris and Ren\'{e} Vidal.
\newblock Theoretical analysis of sparse subspace clustering with missing
  entries.
\newblock In {\em International Conference on Machine Learning}, pages
  4975--4984, 2018.

\bibitem{Tschannen:TIT18}
Michael Tschannen and Helmut B{\"o}lcskei.
\newblock Noisy subspace clustering via matching pursuits.
\newblock {\em IEEE Transactions on Information Theory}, 64(6):4081--4104,
  2018.

\bibitem{Vidal:ACC04}
Ren\'{e} Vidal.
\newblock Identification of {PWARX} hybrid models with unknown and possibly
  different orders.
\newblock In {\em American Control Conference}, pages 547--552, 2004.

\bibitem{Vidal:PRL14}
Ren\'{e} Vidal and Paolo Favaro.
\newblock Low rank subspace clustering {(LRSC)}.
\newblock {\em Pattern Recognition Letters}, 43:47--61, 2014.

\bibitem{Vidal:PAMI05}
Ren\'e Vidal, Yi Ma, and Shankar Sastry.
\newblock {Generalized Principal Component Analysis (GPCA)}.
\newblock {\em {IEEE} Transactions on Pattern Analysis and Machine
  Intelligence}, 27(12):1--15, 2005.

\bibitem{Vidal:Springer16}
Ren\'{e} Vidal, Yi Ma, and Shankar Sastry.
\newblock {\em Generalized Principal Component Analysis}.
\newblock Springer Verlag, 2016.

\bibitem{Vidal:IJCV08}
Ren{\'{e}} Vidal, Roberto Tron, and Richard Hartley.
\newblock Multiframe motion segmentation with missing data using
  {PowerFactorization}, and {GPCA}.
\newblock {\em International Journal of Computer Vision}, 79(1):85--105, 2008.

\bibitem{vonLuxburg:StatComp2007}
Ulrike von Luxburg.
\newblock A tutorial on spectral clustering.
\newblock {\em Statistics and Computing}, 17(4):395--416, 2007.

\bibitem{Wang:ICML15}
Yining Wang, Yu{-}Xiang Wang, and Aarti Singh.
\newblock A deterministic analysis of noisy sparse subspace clustering for
  dimensionality-reduced data.
\newblock In {\em International Conference on Machine Learning}, pages
  1422--1431, 2015.

\bibitem{Wang:JMLR16}
Yu-Xiang Wang and Huan Xu.
\newblock Noisy sparse subspace clustering.
\newblock {\em Journal of Machine Learning Research}, 17(12):1--41, 2016.

\bibitem{Wang:NIPS13-LRR+SSC}
Yu-Xiang Wang, Huan Xu, and Chenlei Leng.
\newblock Provable subspace clustering: When {LRR} meets {SSC}.
\newblock In {\em Neural Information Processing Systems}, 2013.

\bibitem{Xin:TSP18}
Bo Xin, Yizhou Wang, Wen Gao, and David Wipf.
\newblock Building invariances into sparse subspace clustering.
\newblock {\em IEEE Transactions on Signal Processing}, 66(2):449--462, 2018.

\bibitem{Yang:ECCV16}
Yingzhen Yang, Jiashi Feng, Nebojsa Jojic, Jianchao Yang, and Thomas~S Huang.
\newblock $\ell_0$-sparse subspace clustering.
\newblock In {\em European Conference on Computer Vision}, pages 731--747,
  2016.

\bibitem{You:ECCV18}
Chong You, Chi Li, Daniel~P. Robinson, and Ren\'{e} Vidal.
\newblock A scalable exemplar-based subspace clustering algorithm for
  class-imbalanced data.
\newblock In {\em European Conference on Computer Vision}, 2018.

\bibitem{You:CVPR16-EnSC}
Chong You, Chun-Guang Li, Daniel~P. Robinson, and Ren\'{e} Vidal.
\newblock Oracle based active set algorithm for scalable elastic net subspace
  clustering.
\newblock In {\em {IEEE} Conference on Computer Vision and Pattern
  Recognition}, pages 3928--3937, 2016.

\bibitem{You:CVPR16-SSCOMP}
Chong You, Daniel~P. Robinson, and Ren\'{e} Vidal.
\newblock Scalable sparse subspace clustering by orthogonal matching pursuit.
\newblock In {\em {IEEE} Conference on Computer Vision and Pattern
  Recognition}, pages 3918--3927, 2016.

\bibitem{You:ICML15}
Chong You and Ren\'{e} Vidal.
\newblock Geometric conditions for subspace-sparse recovery.
\newblock In {\em International Conference on Machine Learning}, pages
  1585--1593, 2015.

\bibitem{Zhang:CVPR19}
Junjian Zhang, Chun-Guang Li, Chong You, Xianbiao Qi, Honggang Zhang, Jun Guo,
  and Zhouchen Lin.
\newblock Self-supervised convolutional subspace clustering network.
\newblock In {\em Proceedings of the IEEE Conference on Computer Vision and
  Pattern Recognition}, pages 5473--5482, 2019.

\bibitem{Zhang:ICML19}
Tong Zhang, Pan Ji, Mehrtash Harandi, Huang Wenbing, and Hongdong Li.
\newblock Neural collaborative subspace clustering.
\newblock In {\em ICML}, 2019.

\bibitem{Zhang:JMLR19}
Zhenyue Zhang and Yuqing Xia.
\newblock Minimal sample subspace learning: Theory and algorithms.
\newblock {\em arXiv preprint arXiv:1907.06032}, 2019.

\end{thebibliography}
	}

\end{document}